\documentclass[journal, twoside]{IEEEtran}
\let\labelindent\relax
\usepackage{graphicx}
\usepackage{subcaption}
\usepackage{eqlist}
\usepackage{amsfonts}
\usepackage{tabularx}
\usepackage{booktabs}
\usepackage{amssymb}
\usepackage{amsmath}
\usepackage{enumitem}
\usepackage{mathtools}
\usepackage{threeparttable}
\usepackage[lined, ruled, linesnumbered, commentsnumbered]{algorithm2e}
\usepackage{lipsum}

\usepackage[usenames,dvipsnames]{xcolor}
\usepackage{comment}
\usepackage{wrapfig}
\usepackage{soul}
\usepackage{supertabular}
\usepackage{longtable}

\usepackage{multirow}

\usepackage{tikz}
\usetikzlibrary{fit,shapes.misc}

\usepackage{CJKutf8}
\definecolor{bleudefrance}{rgb}{0.19, 0.55, 0.91}
\definecolor{awesome}{rgb}{1.0, 0.13, 0.32}

\definecolor{darkgreen}{rgb}{0.0, 0.65, 0.0}

\begin{document}
\title{Integrating a Manual Pipette into a Collaborative Robot Manipulator for Flexible Liquid Dispensing}
\author{Junbo Zhang$^{1}$, Weiwei Wan$^{1*}$, Nobuyuki Tanaka$^{2}$, Miki Fujita$^2$ and Kensuke Harada$^{1}$
\thanks{$^{1}$Graduate School of Engineering Science, Osaka University, Japan.}%
\thanks{$^{2}$RIKEN, Japan.}%
\thanks{Contact: Weiwei Wan, {\tt\small wan@sys.es.osaka-u.ac.jp}}
}

\markboth{Journal Submission in Review, 2022}
{Zhang \MakeLowercase{\textit{et al.}}: Integrating a Manual Pipette into a Collaborative Robot Manipulator for Flexible Liquid Dispensing} 
\maketitle

\begin{abstract}
This paper presents a system integration approach for a 6-DoF (Degree of Freedom) collaborative robot to operate a pipette for liquid dispensing. Its technical development is three-fold. First, we designed an end-effector for holding and triggering manual pipettes. Second, we took advantage of a collaborative robot to recognize labware poses and planned robotic motion based on the recognized poses. Third, we developed vision-based classifiers to predict and correct positioning errors and thus precisely attached pipettes to disposable tips. Through experiments and analysis, we confirmed that the developed system, especially the planning and visual recognition methods, could help secure high-precision and flexible liquid dispensing. The developed system is suitable for low-frequency, high-repetition biochemical liquid dispensing tasks. We expect it to promote the deployment of collaborative robots for laboratory automation and thus improve the experimental efficiency without significantly customizing a laboratory environment.
\end{abstract}
\def\abstractname{Note to Practitioners}
\begin{abstract}
The system presented in the paper helps to promote low-frequency, high-repetition biochemical experiments. Such experiments were difficult to be performed automatically. For one thing, they are temporal. It is undesirable to spare limited lab space for special-purpose machines. For the other, such experiments require a repeated operation. Employing humans to carry out the experiments is difficult. The integrated robotic manipulator presented in this paper solves the problem using a vertical articulated robot. With the support of the proposed planning and error correction methods, the robot can be placed in narrow lab space and set up quickly for low-frequency, high-repetition experiments. Also, it may be set up together with existing automation devices for simultaneous liquid dispensing and examination. Interested practitioners may find an example of the robot working with a plant phenotyping system for screening chemicals in the supplementary video.
\end{abstract}


\section{Introduction}
\label{sec:introduction}

\IEEEPARstart{T}{his} paper proposes a system integration approach for a 6-DoF (Degree of Freedom) collaborative robot to operate a manual pipette for biomedical liquid dispensing. Unlike methods in existing robotic pipetting systems \cite{fleischer2016application}\cite{fleischer18}\cite{reed21}, the proposed approach does not have any special requirements or need any modification to the experimental environment and labware. It takes advantages of a collaborative robot's ``direct teaching'' mode to recognize initial object poses, employs motion planning \cite{wan2021ras} to generate robot motion online, and leverages modern classification methods \cite{knn2021}\cite{smola2004tutorial}\cite{he2016deep}\cite{dosovitskiy2021an} to precisely attach disposable tips to the pipette. Together with specially designed fingertips and hand palm structures for holding and triggering manual pipettes, the method exhibits high flexibility for using a robot manipulator to carry out liquid dispensing tasks. It is expected to improve the experimental efficiency in a crowded laboratory environment without much customization.

\begin{figure}[t]
    \centering
    \includegraphics[width=.97\linewidth]{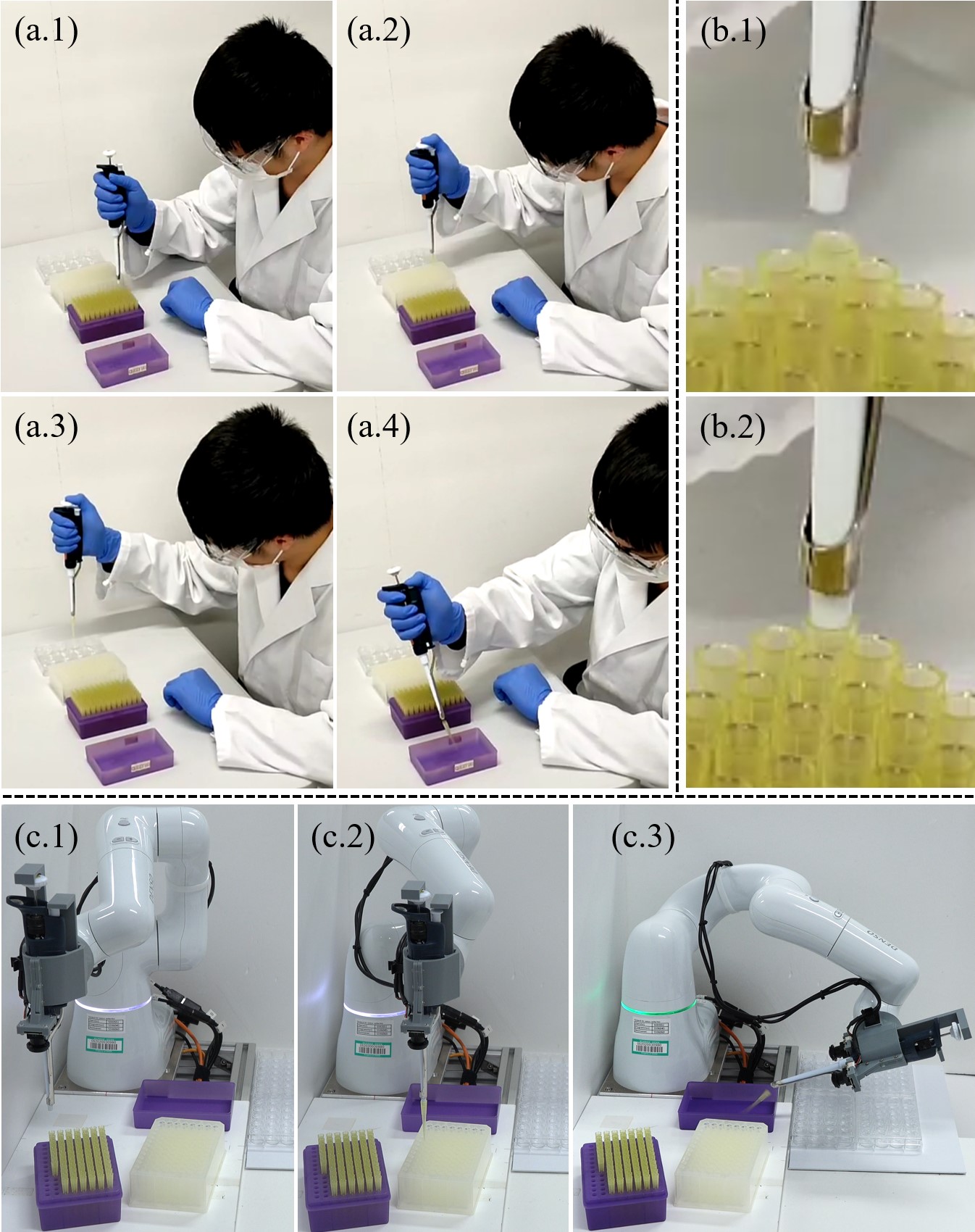}
    \caption{The paper integrates a manual pipette with a 6-DoF collaborative robot for dispensing liquid like a human scientist. (a) A human scientist uses a manual pipette to dispense liquid. (b) Attaching tips need precise control. (c) The robotic manipulation system can perform a similar liquid dispensing task with satisfying precision. The developed system is beneficial because: (i) We take advantage of a collaborative robot's ``direct teaching'' mode to recognize initial object poses. (ii) We designed new fingertips and palm structures to hold and use a manual pipette. (iii) A vision transformer is used to recognize the alignment of the pipette shaft and tip and thus secure robust attachment. Compared to conventional liquid handling robots, the developed system can fit narrow lab spaces and the human environment without much customization.}
    \label{fig:motivation}
\end{figure}

Biochemical liquid dispensing requires much carefulness and patience. Fig. \ref{fig:motivation}(a) exemplifies a liquid dispensing process. The human technician used a pipette to attach disposable tips from a rack in (a.1). Then, he aspired liquid from a micro-plate well in (a.2). After that, he dispensed the liquid to a cultivation plate well in (a.3). Finally, he disposed of the tip into a waste box in (a.4). The process requires careful position control for tip attachment (precise insertion) and much repetition, as shown in Fig. (b.1) and (b.2). Using a robot to replace humans will significantly promote such tasks' efficiency, saving labor costs and lab resources.

Conventionally, liquid handling robots had a gantry structure \cite{kong2012}. They required various labware to be placed inside the gantry frame. It was challenging to fit in a narrow laboratory environment or work with existing automation devices\footnote{Some labs use conveyor belts or circulation chains to move plates or pots. Integrating a gantry robot to work aside them is difficult. Others may be too crowded to fit a gantry robot's bulky body.}. On the other hand, the recently developed manipulator-based systems for automatic biochemical experiments can reach 6D poses \cite{yachie2017robotic}. However, they need to be installed in a robot safety fence or room to get separated from the human environment. It is difficult for them to be deployed in a crowded laboratory and work side-by-side with human technicians or scientists. 

Under this background, we develop a more flexible liquid dispensing system by integrating a collaborative robot. The system leverages collaborative teaching to recognize labware, motion planning to generate robotic motion, and deep-learned visual recognition to obtain robust tip attachment. Our main contribution is three-fold. First, we modified a commercial parallel gripper for handling manual pipettes and hosting RGB cameras. The design provides a low-cost choice for a single robotic arm with a two-finger parallel gripper to use a manual pipette. Second, we take advantage of a collaborative robot's ``direct teaching'' mode to recognize labware and carry out online motion planning based on the recognition results. The recognition and motion planning allows flexibly changing labware poses and types without reprogramming the robot. Third, we formulate tip attachment as a peg-in-hole task and use vision-based classifiers to detect and correct deviations. The peg-in-hole method exploits plastic and unfixed tips to improve system robustness. We carefully designed the methods for collecting training data for the classifiers.

In the experimental section, we compared various vision-based classifiers and their performance in real-world executions. We examined the necessity of using two cameras in the end-effector design and analyzed the influence of direct teaching. We also discussed and examined using previously predicted errors to close the loop and improve system performance.

The remaining part of this paper is organized as follows. Section II presents related work. Section III presents the assumed 6-DoF articulated robot and the end-effector design. Section IV presents direct teaching and motion planning. Section V presents the classification problem and the related data collection and workflow for deviation correction. Section VI presents experiments and analysis. Section VII draws conclusions and discusses future work.

\section{Related Work}

Our related work includes three parts. First, we review laboratory automation, with a focus on robots that handles pipettes. Second, we review robotic peg-in-hole methods. Third, we briefly mention integral motion planning, visual servoing, and control for multi-joint manipulators.

\subsection{Laboratory Automation and Robotic Pipetting}

Laboratory automation has a long history and involves the development of equipment in a broad range of fields. Interesting reviews of laboratory automation history could be found in \cite{sasaki1998}\cite{olsen2012}. In this paper, we zoom in on an important topic of laboratory automation, namely the automation for liquid dispensing and analytical measurements \cite{fleischer17book}. Especially, we focus on and review related studies in robotic liquid handling using pipettes. The topic has a long research record. Kong et al. \cite{kong2012} presented a summary of the syringe or pipette-based robotic liquid handling systems before 2012. For more recent studies, Gerber et al. \cite{gerber2017liquid} used Lego blocks to develop a gantry robot with linear rails to operate syringes and handle liquid. Chory et al. \cite{chory2021enabling} developed the Pyhamilton platform to control liquid-handling robots and multi-channel head pipettes for high-throughput tasks like analyzing population dynamics. Gome et al. \cite{gome2019openlh} developed the OpenLH system based on a uFactory 4DoF desktop manipulator. The manipulator is essentially an xyz linear robot plus a rotating head. Fai{\~n}a et al. \cite{faina2020evobot} developed a 3D printed liquid dispensing system. The system had interchangeable electronic syringe heads for adjusting liquid volumes. Gervasi et al. \cite{gervasi2021open} carefully discussed the design of an electronic syringe head for micro-pumping. Dettigner et al. \cite{dettinger2021open} developed a 5-bar parallel robot for pipetting multi-well plates. The robot used fixed needles as the dispensing head. It is thin and compact and can be fitted to the space between the lens and stage of a microscope for time-lapse imaging. Barthels et al. \cite{barthels20} presented the FINDUS (Fully Integrable Noncommercial Dispensing Utility System). They modified a manual pipette by replacing its pressing mechanism with a screwing one for precise pipetting control.

This paper considers using a six-DoF manipulator for pipetting tasks. In laboratory automation, people used six-DoF manipulators for pick-and-place tasks \cite{sparkes2010towards} instead of pipetting since they were considered less precise \cite{kong2012}. The precision gets even worse if online motion planning is conducted (pose repeatability vs. pose accuracy \cite{Abderrahim06}). Despite the shortage in precision, six-DoF manipulators remain beneficial for pipetting tasks since they have small sizes and high dexterity compared with gantry robots. They could significantly solve the space availability challenge for laboratory automation \cite{genzen2018}. For this reason, we study methods to overcome the six-DoF manipulators' shortage and exploit them for flexible pipetting. We leverage vision-based classifiers for detecting and correcting positioning errors.

\subsection{Motion and Trajectory Planning}

Motion and trajectory planning assures a robot's safe and quick movement. The goal of motion planning is to generate a sequence of collision-free joint angles automatically. Basic techniques of motion planning have been well developed and published as textbooks \cite{choset2005principles}\cite{lavalle2006planning}. Currently, the most popular methods are single-shot probabilistic ones \cite{suarez2018can}. Motion planning suffers from kinodynamic constraints \cite{donald1993kinodynamic}, where the maximum joint velocity and acceleration depend on hardware configurations. Trajectory planning considers time and physical feasibility. It finds the shortest time for a motor to move between two goals while considering the maximum velocity and acceleration constraints. Trajectory planning produces physically feasible motion sequence\cite{constantinescu2000smooth}. 

Previously, motion planning was less discussed in laboratory automation as labware were carefully arranged avoid collision with robots. For trajectory planning, Vandermeulen et al. \cite{vandermeulen2018} presented a time-optimal interpolation method for improving the motion speed of a gantry liquid handling robot. Kuriyama et al. \cite{kuriyama2008trajectory} studied the trajectory planning problem for quickly moving a spoon of liquid while considering spilling avoidance. 

In this paper, we consider both motion and trajectory planning to avoid obstacle regions and, at the same time, move to goals fast. We use RRT-Connect \cite{kuffner00} to plan a collision-free motion, use a pruning post-processor \cite{geraerts07}\cite{hauser2010fast} to create a shortcut path, and use time-optimal path parameterization \cite{pham18} to finally get a fast trajectory. We did not use the optimal planners like Kinodynamic RRT or RRT* \cite{webb2013kinodynamic} since our concern included not only trajectory efficiency but also the planning cost.

\subsection{Robotic Peg-in-Hole}

Tip attachment is essentially a peg-in-hole \cite{broenink1996peg} problem in robotics. A conventional peg-in-hole process can be divided into a search phase and an insertion phase \cite{chen2020jira}. The goal of the search phase is to make the peg overlap with the hole since they may be far from each other in the beginning. Blind search policies \cite{chhatpar01} or visual detection methods \cite{huang13}\cite{kojima18} were widely studied previously to solve the problems in the search phase. The insertion phase, on the other hand, focuses on the peg's penetration into the hole. Policy-based \cite{su2022}\cite{park2017} and control-based \cite{balletti2012towards}\cite{song2016guidance} methods were developed for assuring robust insertion. There were also passive control methods that used compliant mechanisms \cite{haskiya1998passive}\cite{toshihiro17} for adjusting insertion poses.

More recently, studies explored deep learning or deep reinforcement learning to expand the performance and generalization of peg-in-hole methods. The methods could solve the search and insertion phase as a whole. For example, Triyonmoputro et al. \cite{joshua19} used a VGG16 deep neural network trained using synthetic data to align a peg and a hole. The learned results were combined with spiral search and force control to move the peg inside the hole and finish insertion solidly. Yu et al. \cite{yu2019siamese} presented the Siamse convolutional neural network to estimate the pose of an in-hand camera and thus adjust an in-hand male part for precise insertion. Nigro et al. \cite{nigro20} developed a combined detection and control approach where the robot first used a hand-mounted Realsense camera to reconstruct the 3D shape of a workpiece and detect holes. Then, it leveraged force control with joint torque feedback to finish insertion. Schoettler et al. \cite{schoetter20} used images captured from a side-view camera to estimate insertion states and carry out deep reinforcement learning. The rewards were determined using a final state image, connecting signals, and end-effector positions. Lee et al. \cite{lee2020} leveraged multi-modal sensory information and self-supervised learning to predict subsequent actions for peg-in-hole tasks. The authors applied the method to non-circular holes.

This paper uses vision-based deep learning to detect and align a peg and a hole simultaneously. We design a spiral exploration curve and move the robot following the curve to generate the correspondent data between deviated positions and correction vectors. We then use the generated data to train a classifier and learn correction actions in the presence of deviations. Contact feedback and compliant mechanisms are not considered as the pipette and tip are fragile and deformable.


\section{Modifying a Commercial Parallel Gripper for Handling Manual Pipettes}

\subsection{Assumed Robot and Task Requirements}

We first present our assumed robot and task requirements. Note that the method and design are not limited to the specific robot and tasks. They are presented in this early stage to give our readers a concrete understanding of our development. The assumed robot is a COBOTTA collaborative manipulator produced by DENSO WAVE Incorporated. The robot is lightweight and inexpensive. It has a small size to fit a narrow laboratory space. The assumed liquid handling tool is a widely used single-channel Gilson manual pipette. The exemplary task is spreading liquid in the wells of a 96 microplate to four 24-well cultivation plates. The task is also called plate expansion in liquid handling literature \cite{fleischer17book}. The applicable tasks are not limited to this expansion one. In the experiment section, we will show the results of connecting this robot to the Riken Integrated Phenotyping System (RIPPS) \cite{fujita2018ripps} for dispensing chemicals.

\begin{figure}[!htbp]
    \centering
    \includegraphics[width=.95\linewidth]{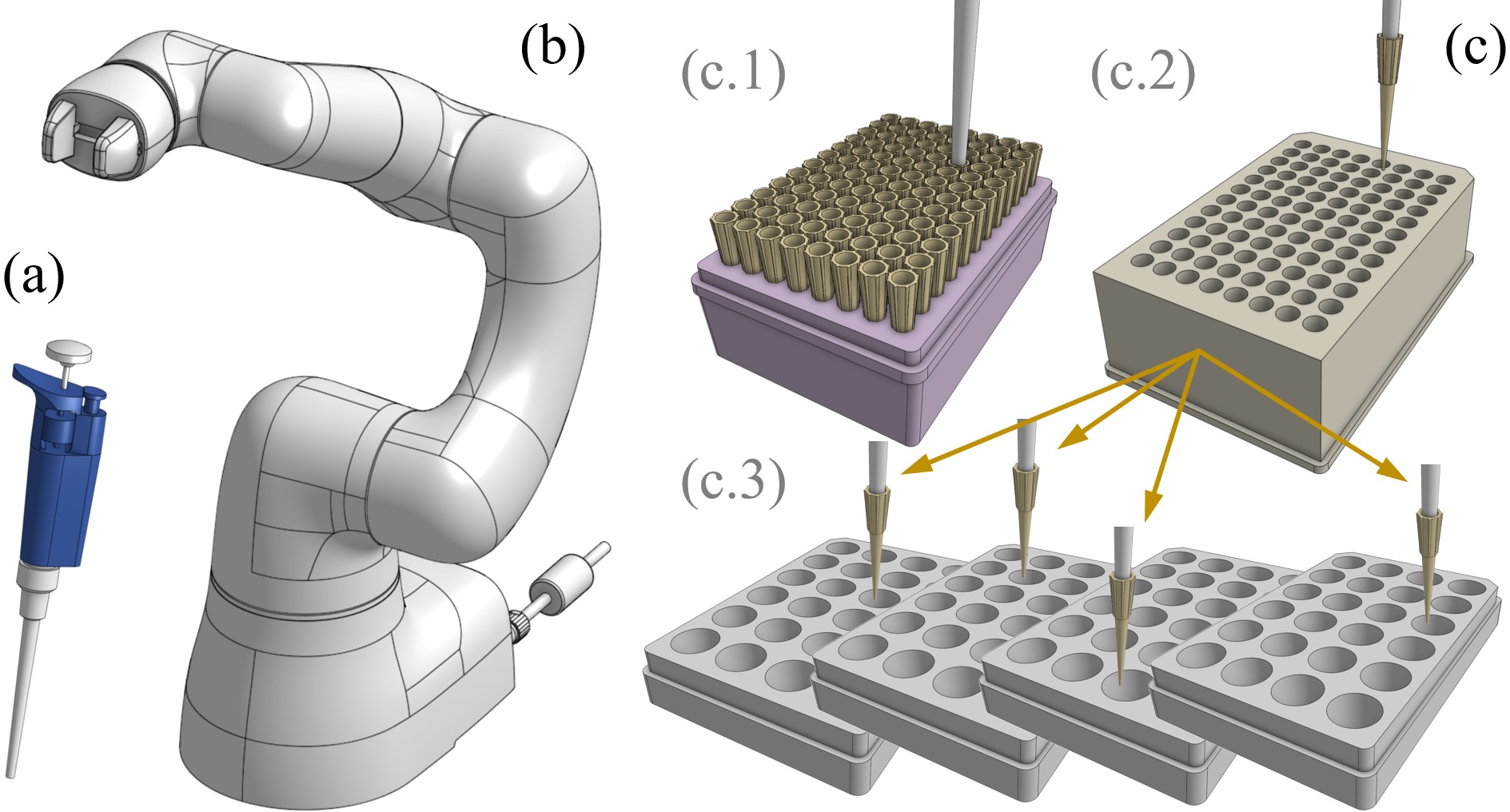}
    \caption{(a) Single-channel Gilson manual pipette. (b) COBOTTA collaborative robot. (c) Labware used in the task and workflow. (c.1) Disposable tips and their hosting rack (96 tips). The pipette is attached to one of the tips. (c.2) 96-well microplate. The pipette is aspiring liquid from one well. (c.3) Expanding liquid to four 24-well cultivation plates.}
    \label{fig:task}
\end{figure}

Fig. \ref{fig:task} shows the pipette, the robot, the disposable tips and their hosting rack (96 tips), and the 96-well microplates and 24-well cultivation plates. The workflow for the given task is illustrated in Fig. \ref{fig:task}(c), where the robot needs to attach a tip from the disposable tip rack (as shown in (c.1)), aspire liquid from a microplate well (as shown in (c.2)), and dispense the liquid to one goal plate well (c.3), and dispose of the tip (not illustrated). The robot must have the following manipulation abilities to perform the task: (1) Hold the pipette. (2) Attach the pipette shaft to a disposable tip. (3) Press the pipette's plunger button to aspire or dispense liquid. (4) Press the pipette's tip ejector button to discard tips. In the following part of this paper, we will present the changes we applied to the commercial Cobotta gripper for holding and triggering pipettes. We will also detail the planning and visual correction methods used to generate motion and achieve robust and flexible liquid dispensing.

\subsection{Gripper Modification}

\begin{figure}[!htbp]
    \centering
    \includegraphics[width=.97\linewidth]{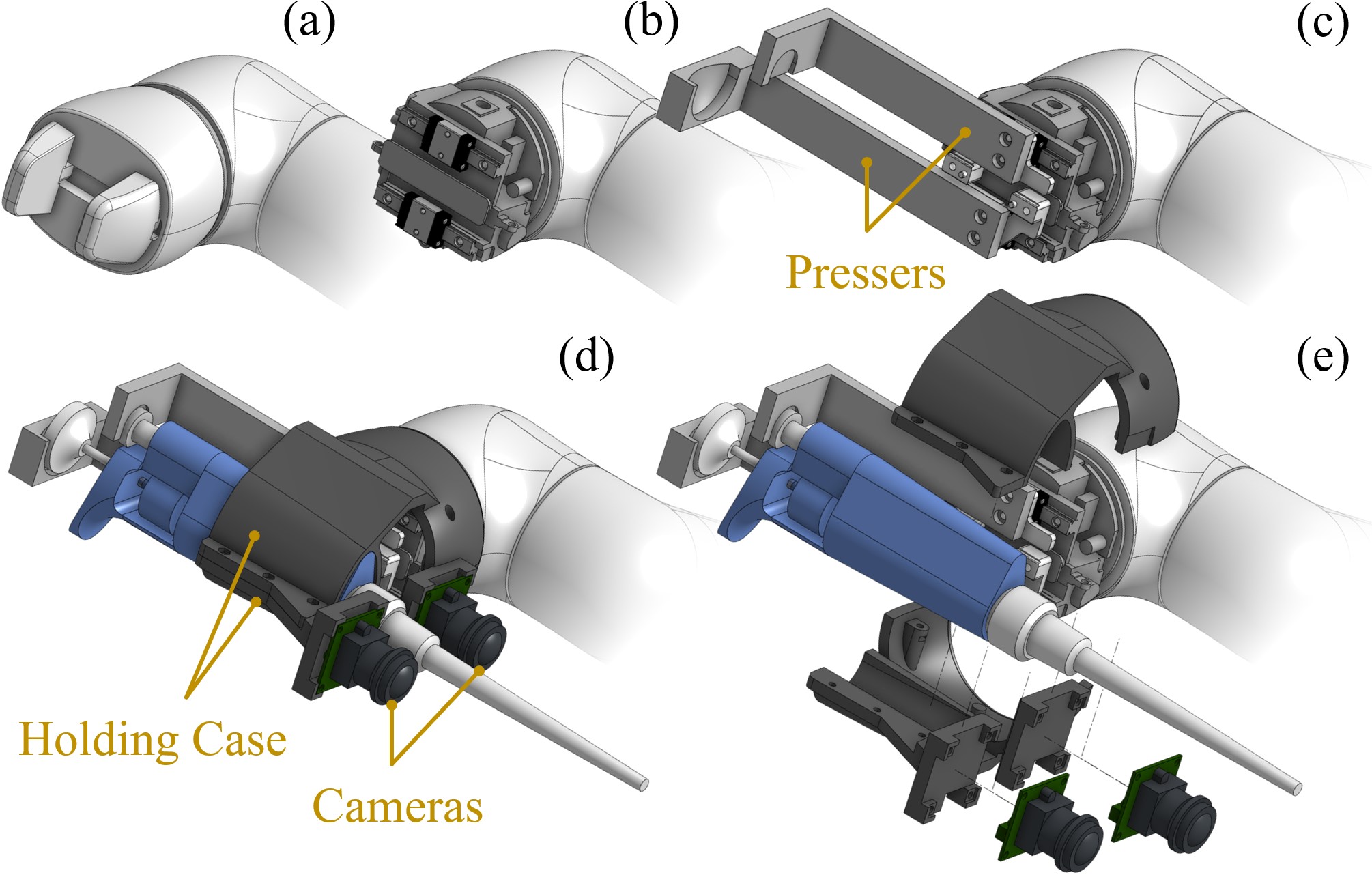}
    \caption{Modifying the commercial Cobotta gripper into a pipetting end-effector. (a) Commercial gripper shipped together with Cobotta. (b) Inner mechanism. (c) L-shape presser fingers. (d, e) Holding case, cameras, and installation method.}
    \label{fig:gripper_modification}
\end{figure}
\begin{wrapfigure}{r}{0.54\linewidth}
  \begin{center}
    \includegraphics[width=\linewidth]{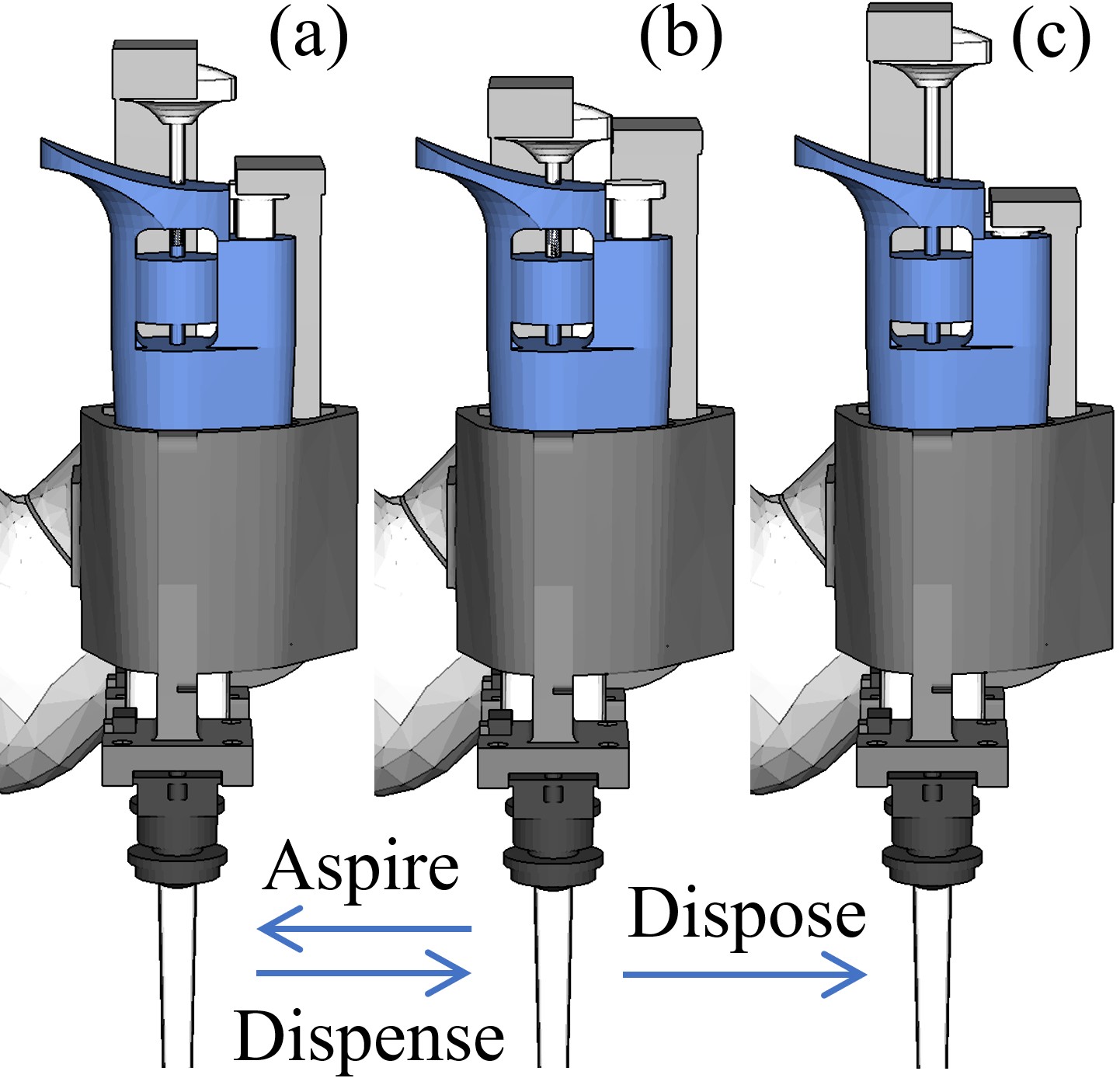}
  \end{center}
  \caption{Using the modified gripper to manipulate the pipette.}
  \label{fig:pipetting}
\end{wrapfigure}

The commercial Cobotta gripper includes a pinion rack mechanism and two linear guides to ensure parallel gripping motion. It cannot simultaneously hold and actuate a pipette. This section's gripper modification aims to adapt the parallel gripper for manipulating pipettes. We do not include a re-development of the electronic systems in the modification. Instead, we use 3D printed cases and fingers to realize the required functions. Fig. \ref{fig:gripper_modification} illustrates the details of gripper modification. Fig. \ref{fig:gripper_modification}(a) is the original commercial Cobotta gripper. The two fingers of the original gripper are attached to two inversely actuated linear guides respectively to realize parallel motion, as shown in Fig. \ref{fig:gripper_modification}(b). In order to manipulate pipettes, we change the two straight fingers of the original gripper to L shapes and use them to trigger the pipette's plunge and tip ejector buttons. Fig. \ref{fig:gripper_modification}(c) illustrates the changed L-shape fingers. They are called pressers in the following context. Fig. \ref{fig:gripper_modification}(d) illustrates the contact between the pressers and the pipette buttons. When the gripper opens, the long presser will press the plunge button, and the short presser will move away from the ejector button, as shown in Fig. \ref{fig:pipetting}(a) to (b). On the other hand, the long presser will release the plunge button, and the ejector presser will move to the ejector button when 
the gripper is closed. Fig. \ref{fig:pipetting}(b) to (c) illustrate the case. Here, we use ``move away from'' and ``move to'' instead of ``pressing'' and `` releasing'' for the ejector button because the shorter presser is not always in contact with the ejector button. It only presses and releases the ejector button in the second half of the gripping stroke. The long presser presses and releases the plunge button in the first half. We take advantage of this difference and open or close the gripper in the first half of the gripping stroke to aspire and dispense liquid. We may also close the gripper to the end to dispose of tips.

Besides the pressers, we design a holding case by performing a Boolean operation against the CAD model of the pipette body. The designed holding case can thus firmly clamp the pipette after being screwed onto the gripper base (Fig. \ref{fig:gripper_modification}(e)). At the bottom of the holding case, we prepare two hosting brackets for attaching cameras. The configuration of the cameras can be seen in Fig. \ref{fig:gripper_modification}(d), (e), and also Fig. \ref{fig:pipetting}. Section V will deliver their details and the related vision methods.

\section{Goal Pose Acquisition and Motion Planning}

In this section, we present methods for obtaining labware poses and generating robot motion.

\subsection{Obtaining Labware Poses with Collaborative Teaching}

\begin{figure}[!htbp]
    \centering
    \includegraphics[width=.97\linewidth]{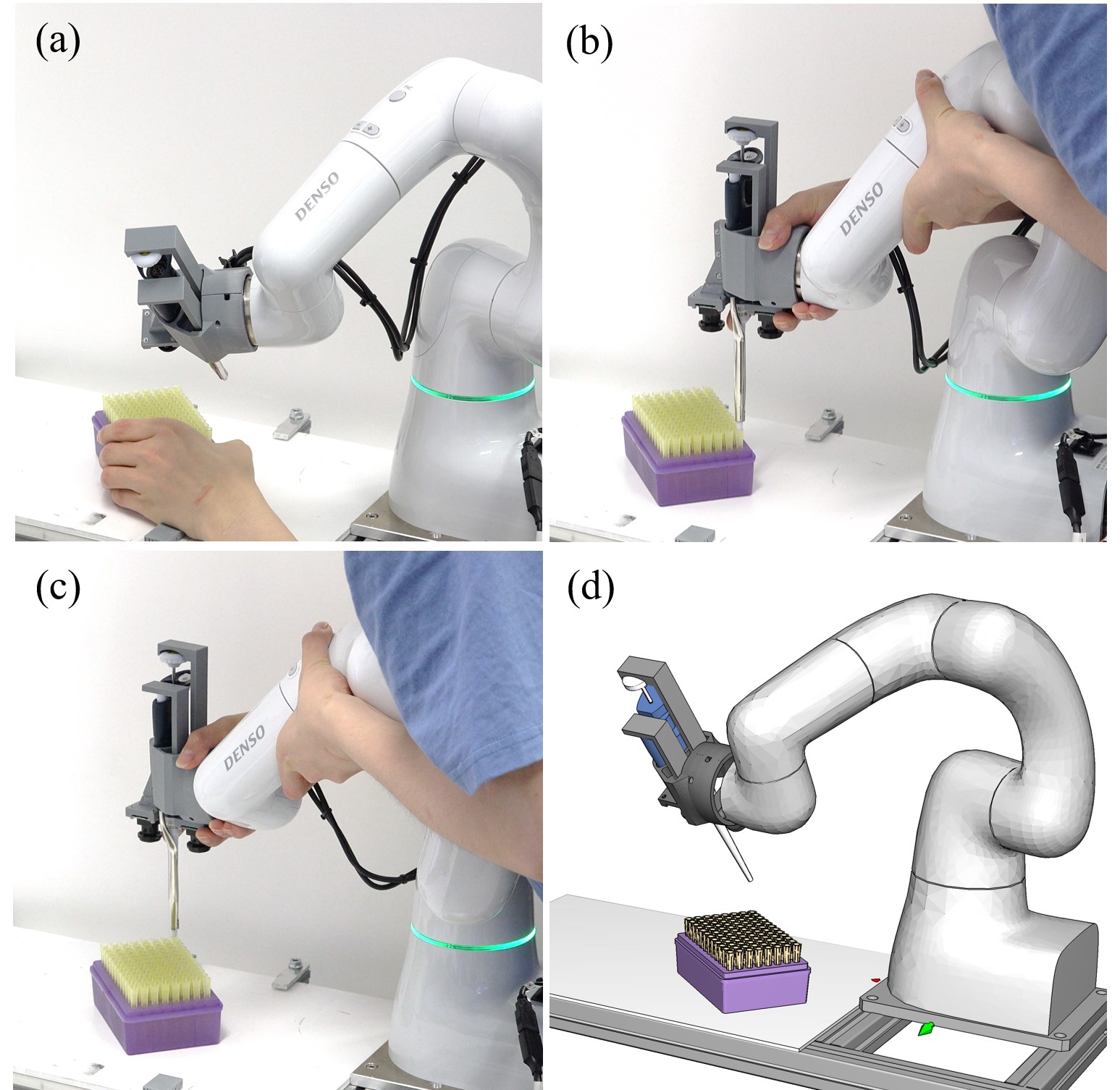}
    \caption{Obtaining the pose of a randomly placed rack. (a) Place down a rack. (b, c) Move the pipetting end-effector to the start and end tip positions of a long edge in the rack to record the corner positions. (d) Obtained rack pose shown in a simulation environment.}
    \label{fig:measure_rack}
\end{figure}

We assume labware to be randomly placed in the workspace of the collaborative robot and take advantage of a collaborative robot's ``direct teaching'' mode to obtain labware poses. Take the tip rack as an example. We move the robot to the starting tip position in the rack and then several other tip positions along the long edge of the rack to record their position values in the robot coordinate system. With the recorded tip position values, we can obtain a pose of the rack in the robot's local coordinate system and compute each tip's coordinates. Fig. \ref{fig:measure_rack} illustrates the process.

The method helps us to avoid complex vision integration. Conventionally, people use a depth camera to detect the tip rack, which requires external calibration between the robot and the global vision system and has difficulty in detecting translucent and crystal objects. Instead of using a depth sensor, we take advantage of a collaborative robot's ``direct teaching'' mode and obtain a rack pose by dragging the robot to several tip positions. Details about the ``direct mode'' could be found in documents of an arbitrary commercial collaborative manipulator. For Cobotta, it is [Cobotta Manual 7273: Switching Between Operation Modes]. 

\subsection{Search for Reachable Goal Poses}

Given a task, the goal of motion planning is to generate a series of robot joint motions that move the pipette across multiple intermediate poses. The first step of motion planning is determining the intermediate goal poses and the robot configurations to reach them. For the plate expansion task mentioned in Fig. \ref{fig:task}, the intermediate goal poses can be simply determined as vertical poses above the tips, the source wells, the goal wells, etc., as illustrated in Fig. \ref{fig:task}(c). However, these poses are not necessarily reachable by the robot. Thus, we develop searching methods to find the reachable intermediate goal poses. 

\begin{figure}[!htbp]
    \centering
    \includegraphics[width=.95\linewidth]{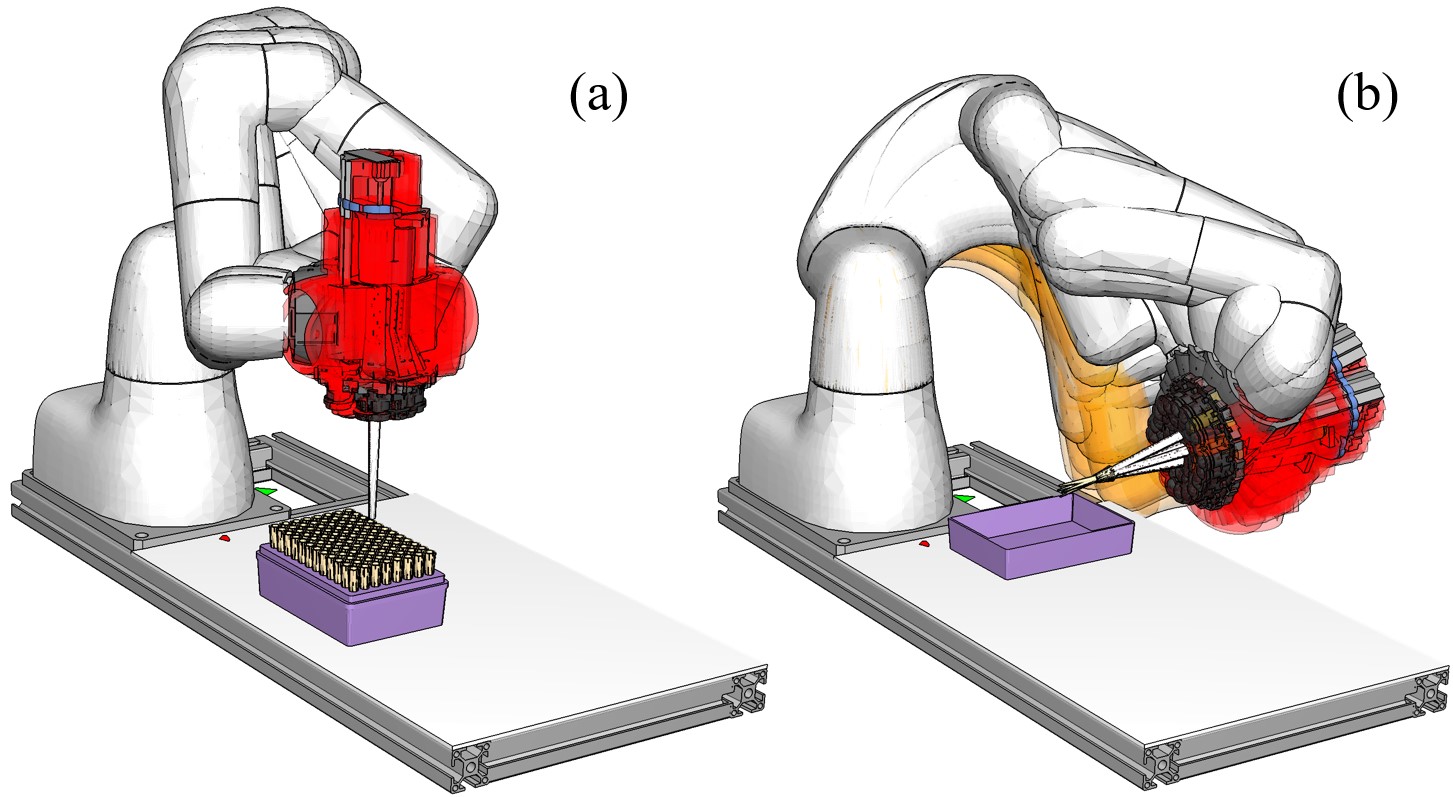}
    \caption{Searching reachable robot configurations for (a) attaching tips and (b) disposing tips. The red end-effector poses are ik-infeasible. The orange robot configurations are in collision. The white robot configurations are the feasible results.}
    \label{fig:nullspace}
\end{figure}

We initially specify seed goal poses for the tip attachment, aspiration, and dispensing key poses as vertical ones above the tips or wells. When a seed goal pose is not reachable, we search with interleaved clockwise and counter-clockwise rotation around the pipette shaft's central axis to find a feasible candidate. Here, ``reachable'' or ``feasible'' mean a goal pose with solvable robotic inverse kinematics. At the same time, the robot at a solved configuration is collision-free. Fig. \ref{fig:nullspace}(a) shows the searching process. We could perform such a search because the pipette shaft and tips have cylindrical mating pairs, and the rotation around the central axis of the pipette shaft can be arbitrary. From the view of a 6-DoF articulated robot, reaching the goal requires 5 DoFs. The robot has one redundant DoF compared to the requirement. The Jacobian matrix spawns a one-dimensional null space, and the intermediate robotic goal pose can be any feasible variation in the space.

The goal poses for disposing of attached tips are different. When a tip is disposed of and dropped into a waste box, it may bounce off the bottom repeatedly. We hope to select a disposing pose that leads to minimum bounces. To carry out the minimization, we assume the bouncing force is caused by the energy stored during impact and can be computed using 
\begin{equation}
    F(t)=\frac{mv(t)}{\Delta t}-\Delta F,
\end{equation}
where $m$ is the mass of the tip, $v(t)$ is the speed when the tip touches the bottom of the waste box, $\Delta t$ is the difference between the moments that the tip touches the bottom and gets bounced off, $\Delta F$ is the lost force caused by adhesion during impact. The speed on touch and the resulted bouncing force are time-relevant. They change along with repeated bounces. The other values are constant. 

Following the equation, we could figure out that lowering $v(t)$ helps to quickly reduce $F(t)$ and thus minimize the times of bounces. Since the first impact, namely $v(t=0)$, depends on the dropping height, we choose the disposing position to be a fixed spot near the waste box. Meanwhile, since the following impacts depend on previous bounce-off heights, we choose the disposing rotation to be inside a tilting cone. One side of the tip will touch the box bottom first when disposed of from the rotation. The bouncing force caused by the touch will rotate the tip around its center of mass, which leads the tip's other side to touch the box bottom without bouncing off too much. The $v(t>0)$ will thus be much smaller, and $F(t)$ will reach 0 soon.
    
We initially specify a seed goal disposing pose like the vertical poses above the tips and wells for attachment and aspiration. If the pose is not reachable, we search with random samples (generated as a spherical sector of an icosphere \cite{wan2016ral}) in the tilting cone to find a feasible candidate. Fig. \ref{fig:nullspace}(b) shows the searching process. As the poses inside the cone have fewer DoFs compared to the robot, the search is also essentially an exploration in the three-dimensional rotational null space of the robot's Jacobian matrix. However, only the movement along the basis vector (the vector corresponding to rotation around the shaft's central line) is completely free. The movement along the other basis vectors is limited to a range to ensure the pipette stays inside the cone. 

\subsection{Path and Trajectory Generation}

After determining the intermediate goal poses, we carry out path and trajectory planning to generate joint space motion. We plan the path using the probabilistic sampling-based method. Possible candidate algorithms include the RRT Connect, RRT*, and their kinodynamic variations. Theoretically, any of the methods could solve our problem. However, considering the high costs of RRT* and the kinodynamic variations, we choose an ad hoc solution. We first use RRT to find a feasible path quickly. Then, we use a pruning post processor to make the path less zig-zag. Finally, we employ the time-optimal path parameterization (TOPPRA) method to make the robot move along the path as fast as possible.

\begin{wrapfigure}{l}{0.55\linewidth}
  \begin{center}
    \includegraphics[width=\linewidth]{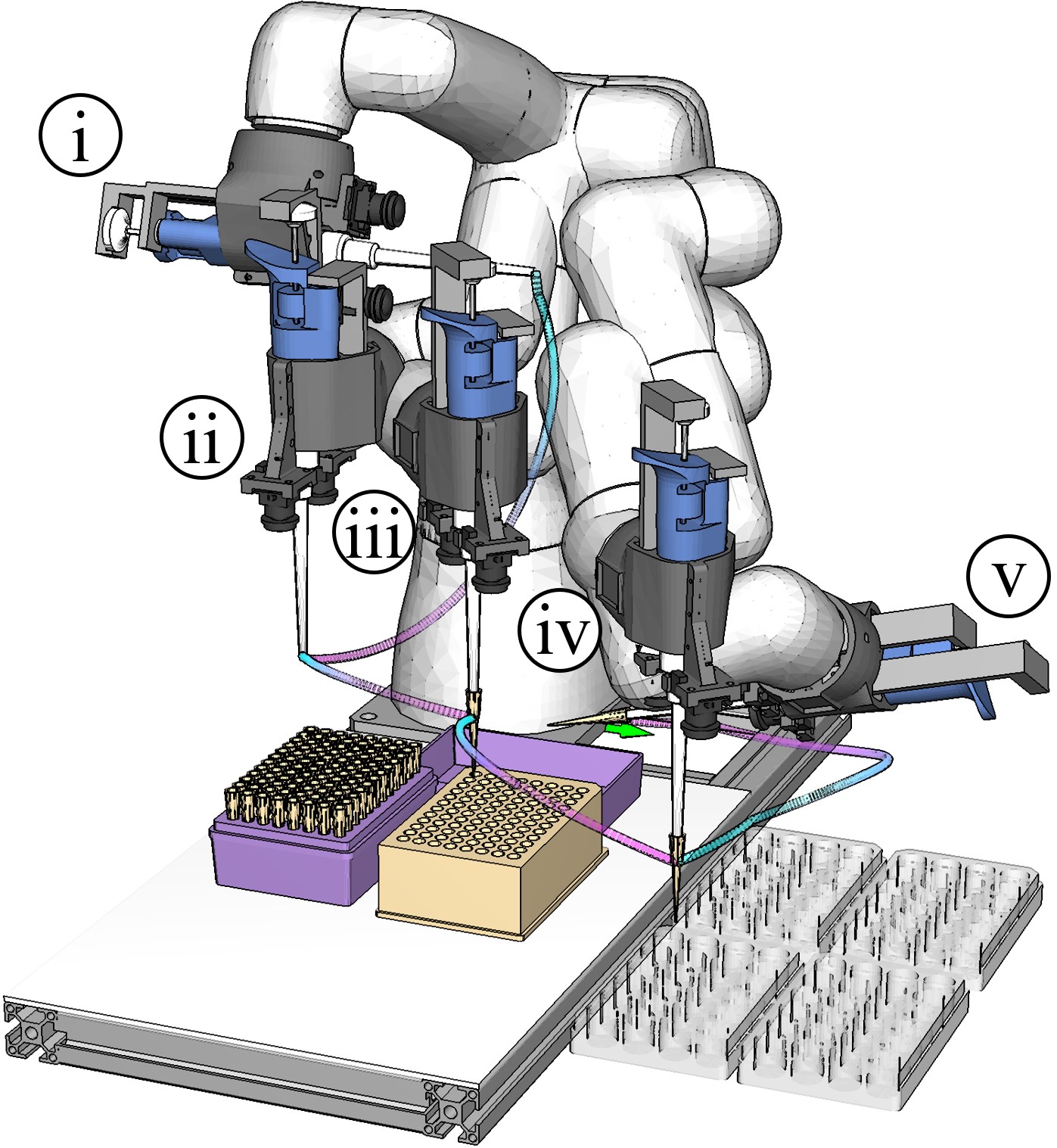}
  \end{center}
  \caption{Using RRT-Connect, a pruning post processor, and TOPPRA to plan fast motion across the intermediate goals (labeled by circled Roman number). The planned trajectories are shown by the cyan-purple curves connecting the pipette tips at the intermediate goals.}
  \label{fig:motionplan}
\end{wrapfigure}

Fig. \ref{fig:motionplan} exemplifies a motion planning result. The intermediate goals in the figure are labeled using a circled Roman number. They include: \textcircled{\raisebox{-0.9pt}{i}} An initial robot configuration. \textcircled{\raisebox{-0.9pt}{ii}} A robot configuration for attaching a tip. \textcircled{\raisebox{-0.9pt}{iii}} A robot configuration for liquid aspiration. \textcircled{\raisebox{-0.9pt}{iv}} A robot configuration for liquid dispensing. \textcircled{\raisebox{-0.9pt}{v}} A robot configuration for disposing the tip. The motion planner generates a motion sequence between every two adjacent goals. Since the motion includes movements of all joints and is difficult to visualize, we only show the trajectories of the tool center point (tip of the pipette) for a rough evaluation. Especially, we illustrate the trajectories using a gradient color, with the cyan end as the start and the purple end as the end.

The motion planning is not necessarily successful. When a planner fails to find a feasible motion, we invalidate the current goal pose pairs and re-explore for a new start and goal combination. We repeat the invalidation and re-exploration until success or final failure (no more reachable goals).

\section{Using Vision-Based Classifiers to Predict and Correct Deviations Before Attaching Tips}
\label{sec:vision}

\subsection{Tip Attachment as a Vision-Based Classification Problem}

The robot suffers from positioning errors at the intermediate goal poses. The reasons are: (1) It is difficult to obtain the precise rack pose through collaborative teaching. (2) The tips may not be strictly at the center of a rack hole. (3) The robot has low absolute accuracy and cannot precisely follow the planned trajectory to reach the planned goal. (4) The previous tip attachment induces disturbances to the rack and makes the rack pose uncertain. (5) Manufacturing defect of the pipette, pipette tips, etc., fabrication and installation errors of the self-designed fingers, holding case, etc. A typical collaborative robot like the Denso COBOTTA cannot compensate for the errors caused by these reasons because the contact forces between the pipette and the tips are small. They require high-quality waist or joint F/T sensors to measure the exact values. The electric current sensors used by most collaborative robots are not precise enough to discern force changes. Also, the tips are not entirely rigid. They are unfixed and partially deformable, making the contact forces change dynamically. Using force control to solve the problem is complex.

\begin{figure}[!htbp]
    \centering
    \includegraphics[width=.97\linewidth]{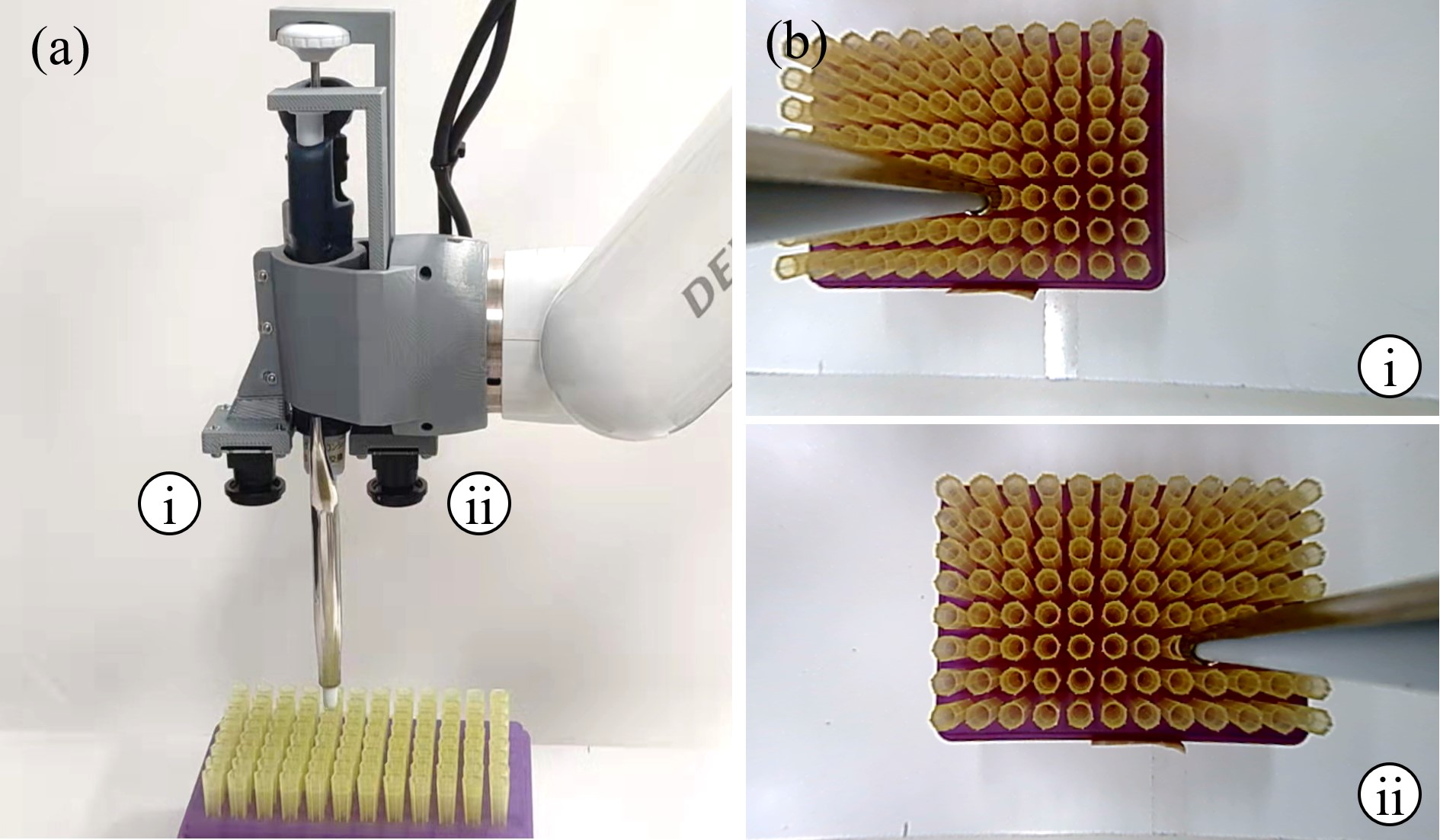}
    \caption{(a) A prototype of the pipetting end-effector. The two cameras for visual detection are labeled by \textcircled{\raisebox{-0.9pt}{i}} and \textcircled{\raisebox{-0.9pt}{ii}}. (b) Images captured by the two cameras.}
    \label{fig:vision}
\end{figure}

Instead of contact and forces, we use vision to align the pipette shaft and the tips in the rack before attachment to ensure successful insertion. We formulate the alignment as a classification problem considering the unfixed and partially deformable tip property. The classifier implementation is open. The only requirement is to accept RGB images collected from cameras and suggest the motion direction and distance needed to align the pipette shaft. Fig. \ref{fig:vision}(a) shows the hardware prototype for our pipetting end-effector, highlighting the vision system. The vision system comprises two RGB cameras for visual detection. Fig. \ref{fig:vision}(b) shows the images captured by the two cameras at the state shown in (a). The classifier must suggest a motion direction and distance for aligning the pipette shaft to the tip near the shaft end in the images for attachment based on the images. 

Consequently, we formulate each class as a deviation from an aligned position. We obtain the classes by starting from an aligned tip position and adding offsets following a spiral path defined in a horizontal plane at the end of the pipette shaft. The origin of the spiral path represents class 0. The pipette shaft aligns with a tip in this class, and the relative position error is 0. Each deviated position along the spiral path represents a different class and has a varying deviation error. We expect a classifier to recognize a pair of captured images as one of the classes so that the system can reduce uncertainty following the classified deviation error. The details will be shown in the following subsection.

\subsection{Spiral Path and Sampling Granularity}

\begin{figure}[!htbp]
    \centering
    \includegraphics[width=.97\linewidth]{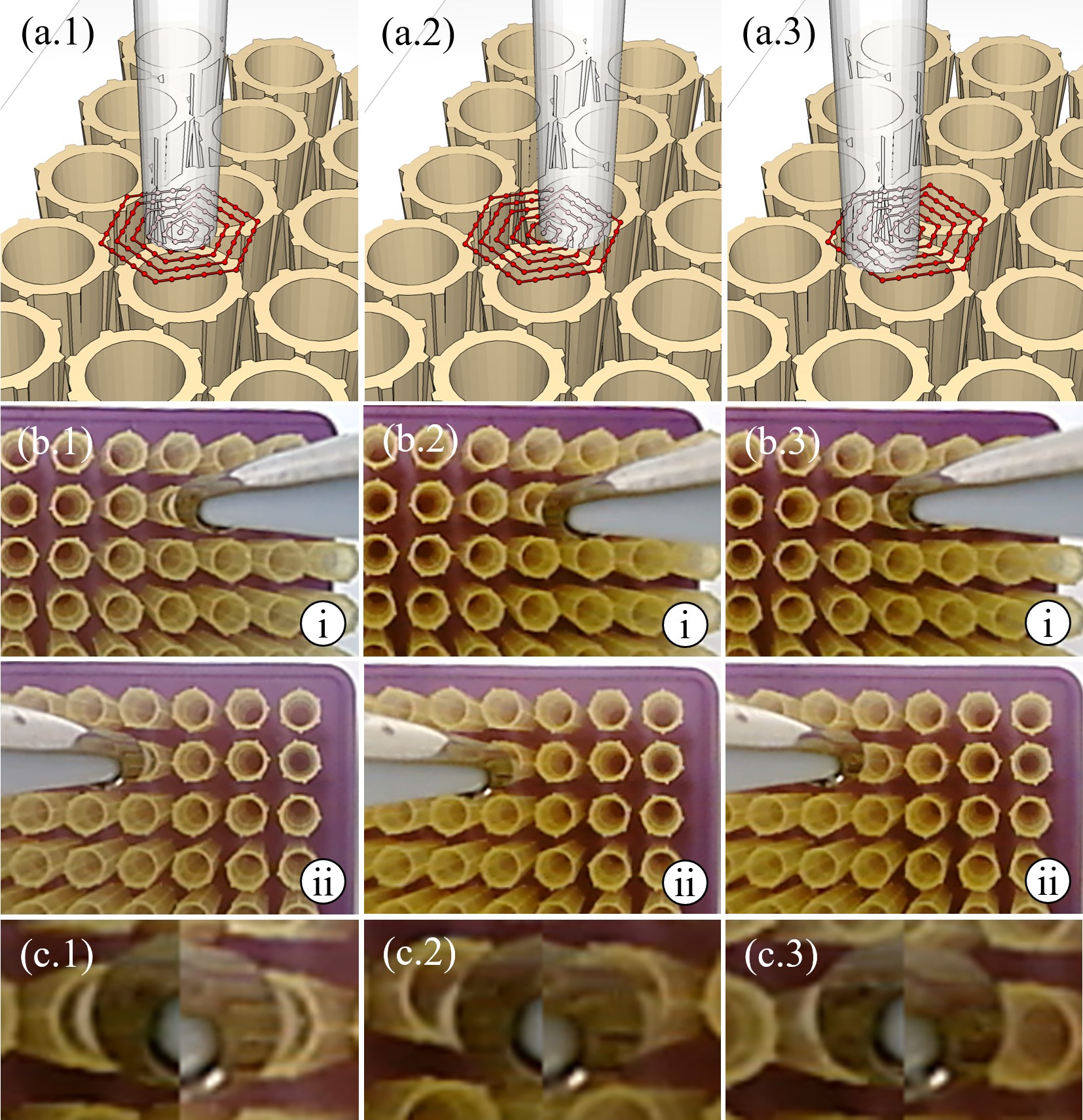}
    \caption{(a) Moving the pipette shaft along the red spiral curve to collect training images. The shaft is at the 0th, 9th, and 29th nodes on the curve in (a.1), (a.2), and (a.3), respectively. (b) Images captured by cameras \textcircled{\raisebox{-0.9pt}{i}} and \textcircled{\raisebox{-0.9pt}{ii}} at the nodes in (a.1-3). (c) Regions near the shaft tip of each captured image pair are cropped and merged into a 45$\times$80 image.}
    \label{fig:cd}
\end{figure}

We obtain the deviation classes by starting from an aligned tip position and adding offsets following a spiral path. Fig. \ref{fig:cd} illustrates the idea. The white cylinders in Fig. \ref{fig:cd}(a) represent the pipette shaft. The yellow objects are the tips. The red lines near the shaft tip are the spiral curve. The spiral curve comprises equally distributed nodes that represent the different deviation classes. The pipette shaft will be moved from the spiral center to each node on the spiral curve sequentially to collect training data for each class. Fig. \ref{fig:cd}(a.1-3) exemplify a movement. The shaft in Fig. \ref{fig:cd}(a.1) is at the spiral center. The shafts in Fig. \ref{fig:cd}(a.2) and Fig. \ref{fig:cd}(a.3) are at the ninth and twenty-ninth nodes respectively. Fig. \ref{fig:cd}(b.1-3) show the images obtained by the two cameras at the nodes in Fig. \ref{fig:cd}(a.1-3) respectively. We crop and merge them into the 45$\times$80 images shown in Fig. \ref{fig:cd}(c.1-3) to train classifiers. Given a pair of newly observed images, we will crop and merge them and use the classifier to recognize the merged image as one of the deviation classes. Then, the system will determine how to adjust the pipette shaft based on the classification result.

\begin{wrapfigure}{r}{0.52\linewidth}
  \begin{center}
    \includegraphics[width=\linewidth]{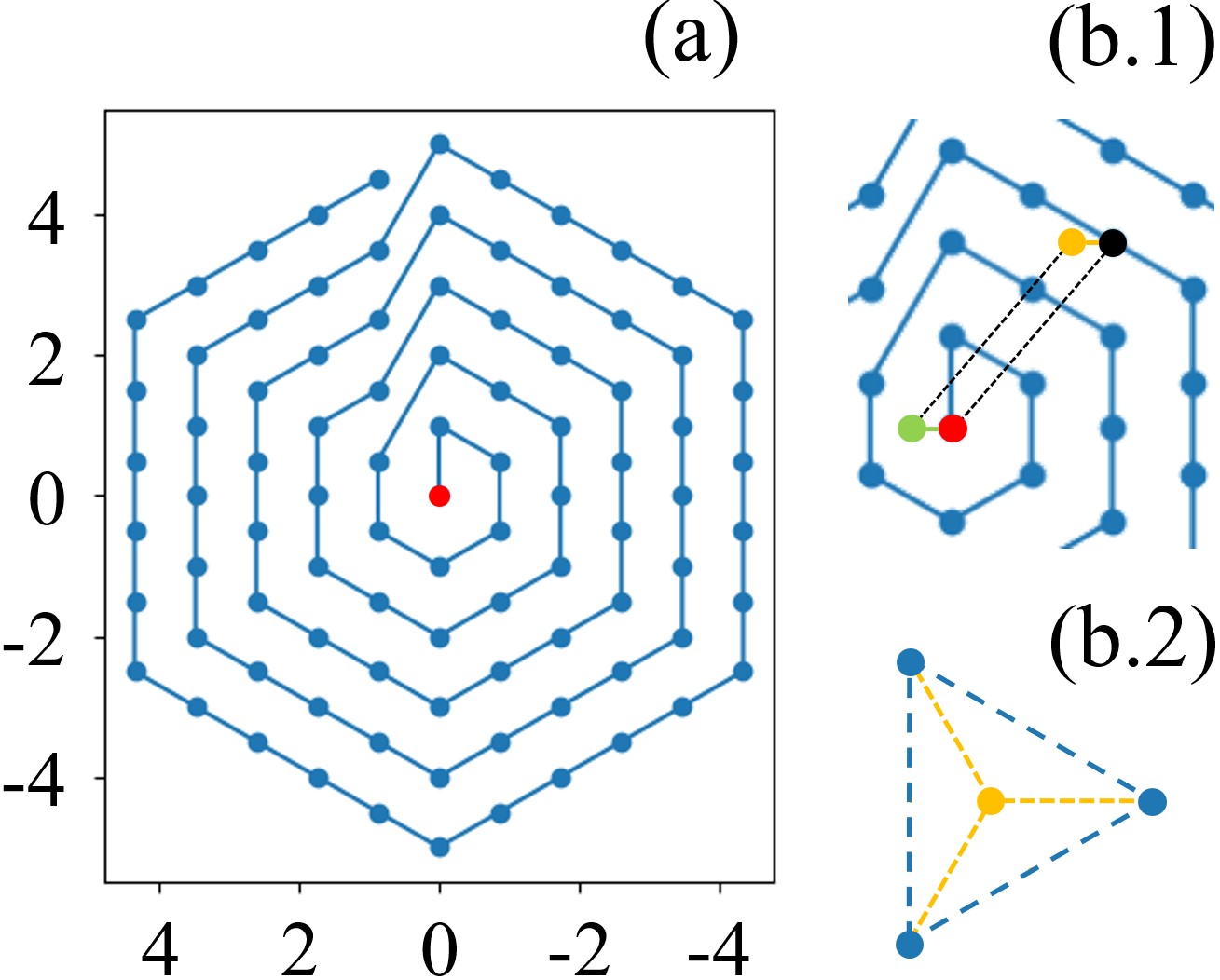}
  \end{center}
  \caption{(a) A 2D view of the proposed spiral curve. (b.1) An uncertain position (yellow) is classified as the class represented by the black node. It is corrected to the green position following the relation between the black node and class 0 (red node). There remains a residual error after correction. (b.2) Maximum residual error appears when the uncertain position is at the center of an equilateral triangle.}
  \label{fig:eql}
\end{wrapfigure}

Fig. \ref{fig:eql}(a) shows a clearer view of the spiral curve using a 2D sketch. We designed the curve by concentrically stacking a series of concentric regular hexagons. We rotated, scaled, and sampled the hexagons so that the distances between two nearby sampled nodes (blue nodes in the figure) have the same value. The 2D space is thus equally triangulated by the sampled nodes starting from the concentric center. Given the camera images obtained at a deviated pipette position, we
train a classifier to classify them to a class represented by a sampled node. The system can then correct the deviation by moving the pipette following the direction and distance identified by the vector connecting the node and concentric center. Fig. \ref{fig:eql}(b.1) illustrates the classification and correction process. The red node is the concentric center. The yellow point denotes a deviated pipette position. The yellow point is classified as the deviation class represented by the black node. The system will correct the deviation by moving the pipette following the direction and distance identified by the vector that connects the black and red nodes. The green point shows the pipette position after correction.

We use such an equilaterally triangulated spiral curve because the nodes are even and provide convenience for identifying the largest residual error after correcting the deviation. The classification-based method has an inherent problem that the corrected pipette position remains to have a residual error from the goal. The residual error is the same as the offset between the original pipette position and the node it was classified to. Take Fig. \ref{fig:eql}(b.1) for example. After being moved to the green position to reduce uncertainty, displacement remains between the green point and the red node. The displacement is the residual error. Its value equals the displacement between the yellow point and the black node. In practice, the residual error must be small enough to insert the pipette into a tip successfully. Thus, we use a tip's acceptable residual error to determine the spiral curve's equilateral triangulation distance\footnote{The tips are plastic and are not tightly fixed on the rack. They provide certain compliance to tolerate the residual error, making the error not need to be zero.}. See Fig. \ref{fig:eql}(b.2) for instance. Since the distances between two nearby nodes on an equilaterally triangulated spiral curve are equal, the triangle that encloses the original pipette position is regular. The maximum displacement of a pipette position from a node will be $\sqrt{3}/3$ of the edge length (yellow dashed segment in the figure). The maximum displacement should be smaller than or equal to the maximally acceptable residual error (denoted as $e$). The equilateral edge length thus should be smaller or equal to $\sqrt{3}e$.

After determining the edge length for equilateral triangulation, we further obtain the maximum number of concentric regular hexagons that form the spiral curve and also the number of nodes on the curve: Suppose the distance between two tips in a rack is $d$, the maximum number of concentric regular hexagons can be computed as $d/(2\sqrt{3}e)$, and the number of nodes can be computed using $3\times{d/(2\sqrt{3}e)}\times{(d/(2\sqrt{3}e)+1)}+1=3d(d+2\sqrt{3}e)/(12e^2)+1$. This work assumes a 200 $\mu l$ pipette and a rack with 96 tips. The values of $e$ and $d$ are respectively 0.5 mm and 9 mm under the assumption. The maximum number of concentric regular hexagons and the number of nodes are thus, respectively, 5 and 91.

\subsection{Rotational Variation under Fabrication Uncertainty}

We changed the pipetting end-effector's position along the spiral path in the previous subsection. We did not discuss its rotation and assumed the end-effector maintained the same rotational pose. In this subsection, we further discuss changes in rotation.

\begin{figure}[!htbp]
    \centering
    \includegraphics[width=.97\linewidth]{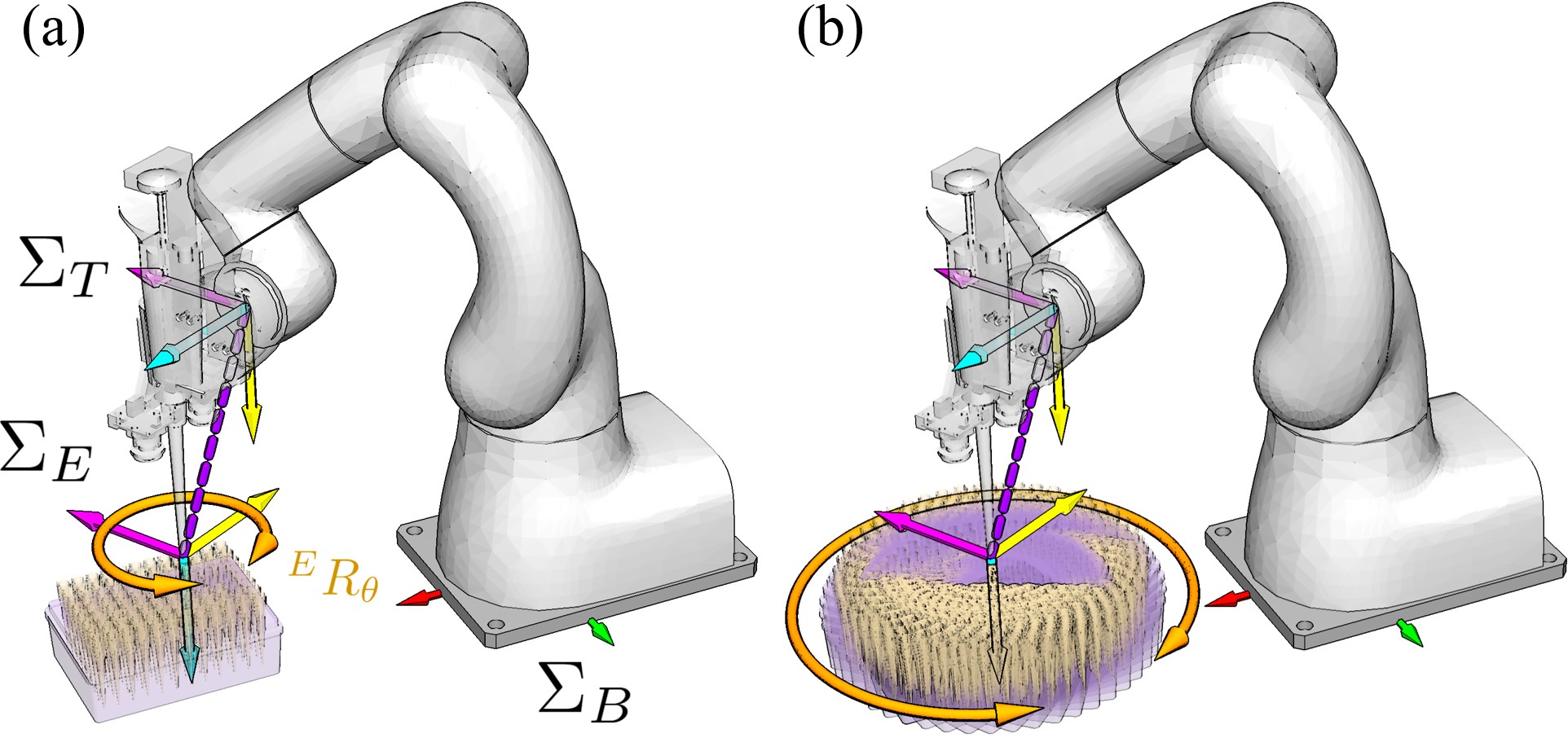}
    \caption{(a) Including rotational variations around the pipette tip. $\Sigma_B$, $\Sigma_T$, and $\Sigma_E$ denote the local coordinate systems at the robot base, the tool mounting flange, and the shaft tip. $\Sigma_E$ is uncertain due to fabrication and installation problems. (b) Avoiding calibrating $\Sigma_E$ by rotating the rack instead of the robot and the pipetting end-effector.}
    \label{fig:rot_tip}
\end{figure}

Fig. \ref{fig:rot_tip}(a) shows the local coordinate systems related to the dispensing robot. Notation $\Sigma_B$ represents the robot manipulator's base coordinate system. Notation $\Sigma_T$ represents the coordinate system at the manipulator's tool mounting flange. Notation $\Sigma_E$ represents the coordinate system at the tip of the pipette shaft. The origins of $\Sigma_B$, $\Sigma_T$, and $\Sigma_E$ are marked as $p_B$, $p_T$, and $p_E$, respectively. At each node on the spiral path, the shaft tip may have various rotations around the local upright axis of $\Sigma_E$ while maintaining an unchanged $p_E$. Unfortunately, it is difficult to apply the rotational variation as the value of $p_E$ is uncertain. The following problems cause uncertainty: (1) We fabricated the end-effector case and fingers using a 3D printer. There exists printing uncertainty; (2) The pipette is clamped by two pieces of 3D printed holding case. Installation uncertainty exists; (3) The exact CAD model of the commercial pipette is not accessible. We used a 3D scanner to reversely engineered it. There exists scanning and modeling uncertainty. These problems make it difficult to measure the exact pose of $\Sigma_E$ and thus the exact position of $p_E$. An intuitive way to solve the problem is to calibrate $\Sigma_E$, which requires a high-precision laser measuring system to detect the center position of the shaft tip. The method is costly and annoyingly needs a re-calibration for each prototype update (re-installation, replacement, etc.) and a new calibration for each fabrication. 

\begin{figure*}[!htbp]
    \centering
    \includegraphics[width=.97\linewidth]{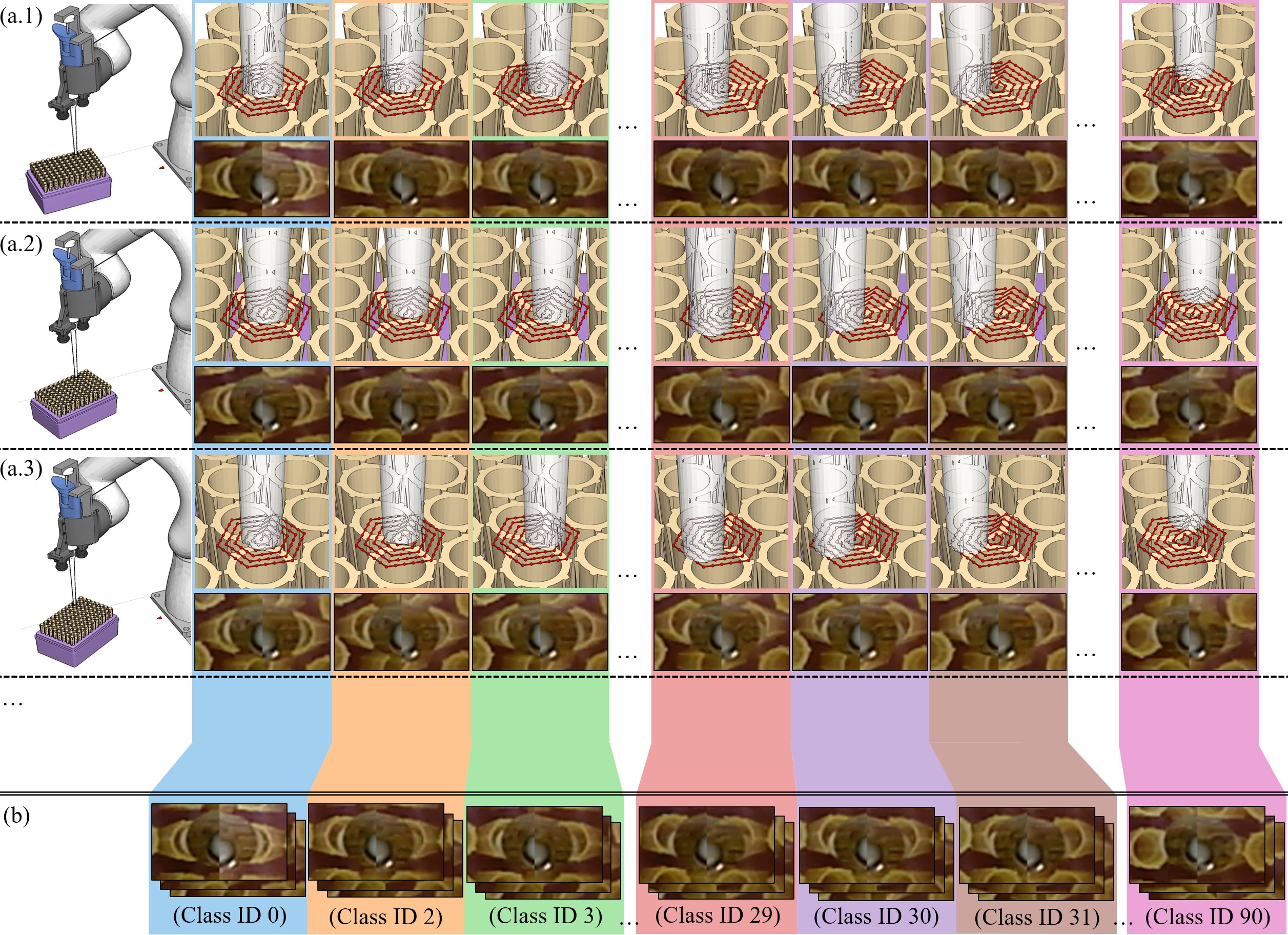}
    \caption{Collecting images considering different rack rotations. The shaft tip of the pipette end-effector is at the same node for each column, but the rack is rotated to different angles. (a.1-3) illustrate different rack rotations. After including the rotation, there will be a collection of images corresponding to each node. (b) shows the image collections. The image collections and their corresponding node classes will be used to train classifiers.}
    \label{fig:rot_rack}
\end{figure*}

Thus, instead of explicitly carrying out a calibration, we propose a calibration-free solution by rotating the tip rack, as shown in Fig. \ref{fig:rot_tip}(b). Rotating the tip rack is essentially the same as rotating the pipetting end-effector since the rotation between the pipetting end-effector and the tip rack is relative. Fig. \ref{fig:rot_rack} exemplifies several varying rotations at different spiral nodes and the correspondent image data. In each sub-row of Fig. \ref{fig:rot_rack}(a), the tip rack is rotated around the central axis of a selected tip in a range [$\theta^-$, $\theta^+$] with a $\delta\theta$ interval. Here, we constrain the range and interval values by the rotation limits of the pipetting end-effector and the granularity used in searching for reachable intermediate goal poses. For the specific robot and system design, we set the range value to be [-30$^\circ$, 30$^\circ$] plus [150$^\circ$, 210$^\circ$], and set the interval value to be 5$^\circ$. We design the range value with two sections to account for the symmetric rack shape. At each rotation, the pipetting end-effector is translated along the same spiral path to collect training data. In total, there will be a collection of images for each node. Suppose images captured at an uncertain pipette position are similar to an element in a collection. In that case, the uncertain position will be classified to the collection's corresponding node class. The last row of Fig. \ref{fig:rot_rack} illustrates the image collections and their correspondent class IDs. These image collections and their correspondent ID values will be used to train a classifier for classifying newly captured images.

\subsection{Availability of Tips in a Rack}

\begin{figure}[!h]
    \centering
    \includegraphics[width=.97\linewidth]{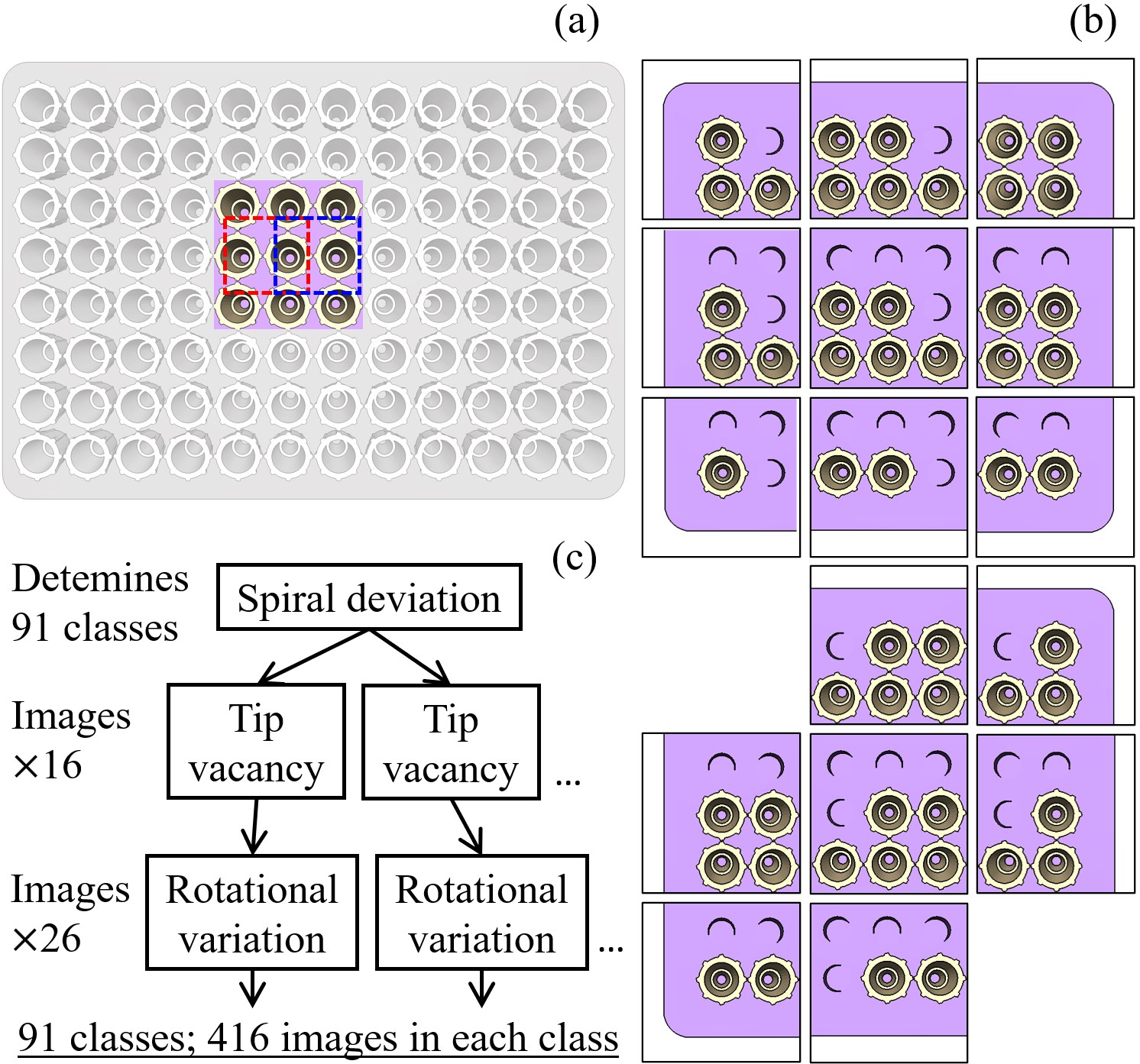}
    \caption{(a) Only the surrounding eight tips influence the merged images. The red and blue dashed boxes show the effective regions for the merge. (b) Tip availability patterns when picking them up one-by-one by starting from a corner tip and moving along the longer edge of the hosting rack. (c) Spiral deviation determines 91 classes. Tip availability and rotational variations increase diversity and expand the training images in each class. In total, we obtain 416 images for each class.}
    \label{fig:vacancy}
\end{figure}

Up to now, we assume all tips are in the rack. The assumption is unreasonable as the robot may have already used and disposed of some tips. We also need to consider rack states with removed tips. Nevertheless, faraway tips are out of sight since the merged images are small. We only need to consider the availability of a target tip's eight surrounding neighbors and remove or keep them during data collection. Fig. \ref{fig:vacancy}(a) exemplifies the above discussion. The colored tips in the figure are the current focus. The pipette is to be aligned with the central one. When considering the availability of the tips, we only need to regard the surrounding eight colored ones, as faraway tips do not influence the merged image. The regions that will be kept in a merged image are highlighted by dashed red and dashed blue frames in the figure. Tips outside the 3$\times$3 tip grid are irrelevant.

The total number of availability combinations for the surrounding eight tips could be $2^8+2^5+2^3 = 296$. The first item in the left-hand side of the equation, namely $2^8$, represents the availability combinations when a central tip is not on edge. The second item, $2^3$, represents the availability combinations when a central tip is at a corner. The third item, $2^5$, represents the availability combinations when a central tip is on edge. The equation did not consider symmetry. After removing symmetric patterns, there would remain 110 possibilities. Collecting data for all of the possibilities is challenging. Luckily we do not need to buckle down on it from a practical viewpoint. The robot picks up the tips one-by-one by starting from a corner tip and moving along the longer edge of the hosting rack. There would be only 16 availability combinations under the picking routine, as shown in Fig. \ref{fig:vacancy}(b).

After considering the tip availability and the rotational variation, we finally have $91$ goal classes for our classification problem. For each class, we obtain 416 training images. The images cover different orientations and tip availability, making the classifier applicable to varying rack poses and tip states on a table surface. The flowchart in Fig. \ref{fig:vacancy}(c) summarizes how we expanded the training images in addition to translational deviation.

\subsection{Exceptions}

\begin{figure}[!h]
    \centering
    \includegraphics[width=.97\linewidth]{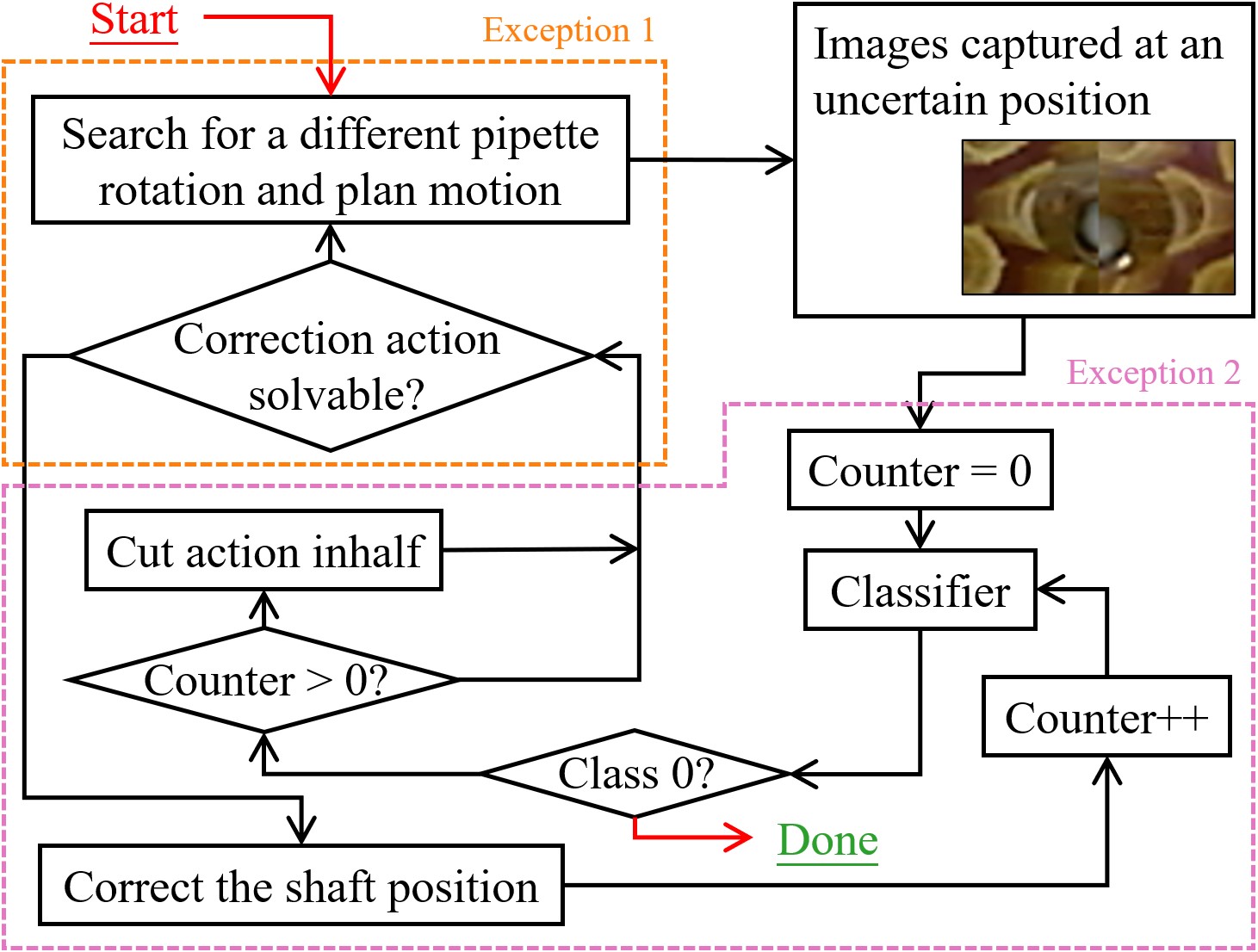}
    \caption{Workflow for using the trained classifier to predict and correct positioning errors. The sections highlighted by the orange and purple dashed boxes were carefully designed to avoid exceptions.}
    \label{fig:exception}
\end{figure}

Fig. \ref{fig:exception} shows the workflow for using the trained classifier to predict and correct positioning errors. Two exceptions need to be carefully resolved when using the classifier. First, although the classifier helps identify a class ID and thus the correcting action, the robot may fail to find a feasible IK or motion to follow the action. When encountering this exception, we follow the same routine presented in Section IV to search for other feasible rotations. The orange dashed box in Fig. \ref{fig:exception} highlights the solution.

Second, we re-run the classifier again after correction to judge if the corrected pose is good enough for insertion (namely, if the corrected pose belongs to class 0). If not, we will repeat the correction process until satisfaction. The repeated correction may result in an infinite correction loop since the corrected pose might never be judged as class 0, and the correction is performed forever. We solve the problem by actively detecting the repetition. When a corrected pose is not judged as class 0, we cut the motion distance for the next correction in half to avoid overshooting. Cutting into halves is helpful because when a corrected pose is wrongly judged, it might be mistaken as a class around 0 (classes 1,2,3,4,5, or 6). The pipetting end-effector will be corrected using the action associated with the wrongly judged class and moved to an opposite side across class 0. If it is again judged as a wrong class, the new wrong class must be diagonal to the previous wrong class across the spiral center. The infinite loops thus will happen between classes 1 and 4, 2 and 5, or 3 and 6. When the motion distance is cut in half, the corrected pose will be held near the spiral center to avoid moving too far to the diagonal side. The purple dashed box in Fig. \ref{fig:exception} highlights the solution.

\section{Experiments and Analysis}

Our experiment section focuses on the vision correction aspect and the systematic liquid dispensing performance. We also briefly mention a chemical dispensing project which involved deploying the developed system to work with an existing automation machine. 

We use the aforementioned COBOTTA collaborative robot from Denso Wave and PIPETMAN Classic P200 pipette from Gilson to build the hardware platform. We use a 96-tip rack from Molecular BioProducts, a 96-deep well microplate from Greiner Bio-One (U-bottom Masterblock\textsuperscript{\tiny\textregistered} storage plate), and four 24-well cell culture plates from As One Cooperation (VTC-P24) to set up the experiment environment. The robot is installed on an aluminum base. The tips, racks, and disposing tray are assumed to be randomly placed on the base. The goal culture well plates are assumed to be randomly placed beside the base. There are no requirements on the exact positions and rotations of the various objects. The planner generates robot motion online for newly detected objects, and the vision-based classifier helps to ensure successful alignment and robust executions. The vision system comprises two RGB cameras made by ELP (ELP-USB30W02M). Both of them have the 480$\times$680 resolution. These two cameras are used because they are the smallest single-chip ones we could find on an easily accessible market (i.e., Amazon.com). It does not need to meet a strict standard and manufacturing precision.

\subsection{Performance of Various Classifiers}

The classifier for vision correction is open. In this subsection, we present the performance of four representative classifiers implemented in our experiments. The classifiers included: Auto-Encoder with KNN (AE-KNN) \cite{knn2021}, Auto-Encoder with SVM (AE-SVM) \cite{smola2004tutorial}, Resnet50 \cite{he2016deep} and Vision Transformer (ViT) \cite{dosovitskiy2021an}. The first two classification methods are conventional: The KNN method achieves classification by identifying the elements that are near to a query example. It is easy to implement and applies to any data with a similarity measurement; The SVM method is theoretically optimal and competitive with neural network methods. It has few critical parameters and could easier to be achieved with fewer samples. An auto-encoder compresses the original image and extracts features before applying the KNN and SVM methods. The Resnet and ViT methods are based on neural networks. Conventionally, neural network-based visual classification or detection can be divided into two categories. The first category is built on the Convolutional Neural Network (CNN), which convolves nearby pixels in an image with learned weights to determine classes. The second category is built on the attention mechanism, which finds global relations and estimates regional relevance for classification. The Resnet50 and ViT methods represent the two categories, respectively. The parameter settings for the four methods in our implementations are as follows: We used the simplest Euclidean distance for similarity measurement in KNN, and chose the K value to be 5; We used an RBF (Radial Basis Function) kernel for SVM; The auto-encoder for the KNN and SVM had six convolutional and three deconvolutional layers. They accepted the 45$\times$80 images and output a 1$\times$720 vector; The Resnet50 was the standard one available in many modern deep learning libraries. There was no modification on it; For the ViT, we divided the 45$\times$80 images into 5$\times$5 patches and reshaped them into a one-dimensional patch vector as the input. 

We trained the four classifiers using the 91 classes of data (416 images in each class) mentioned in Section \ref{sec:vision}, and examined their performance in correcting abbreviations. The examinations included two parts. In the first part, we compared the difference between predicted deviations and their correspondent ground truth, and examined each classifier's prediction precision. In the second part, we carried out physical attachments and examined the success rates. Details of the two parts are as follows.

\subsubsection{Precision of predicted deviations} In this part, we first manually aligned the pipette shaft and a tip in the rack, and assumed the manually aligned position to be the correction goal. Then, we let the robot move away from the aligned position with many randomly generated deviations. The values of the randomly generated deviations were known as the ground truth. After moving to each deviation, we took pictures with the two cameras, cropped and merged them, and used the different classifiers to judge the merged images into deviation classes. We compared the difference between the predicted deviations and their correspondent ground truth to understand each classifier's performance.

\begin{figure}[!htbp]
    \centering
    \includegraphics[width=.97\linewidth]{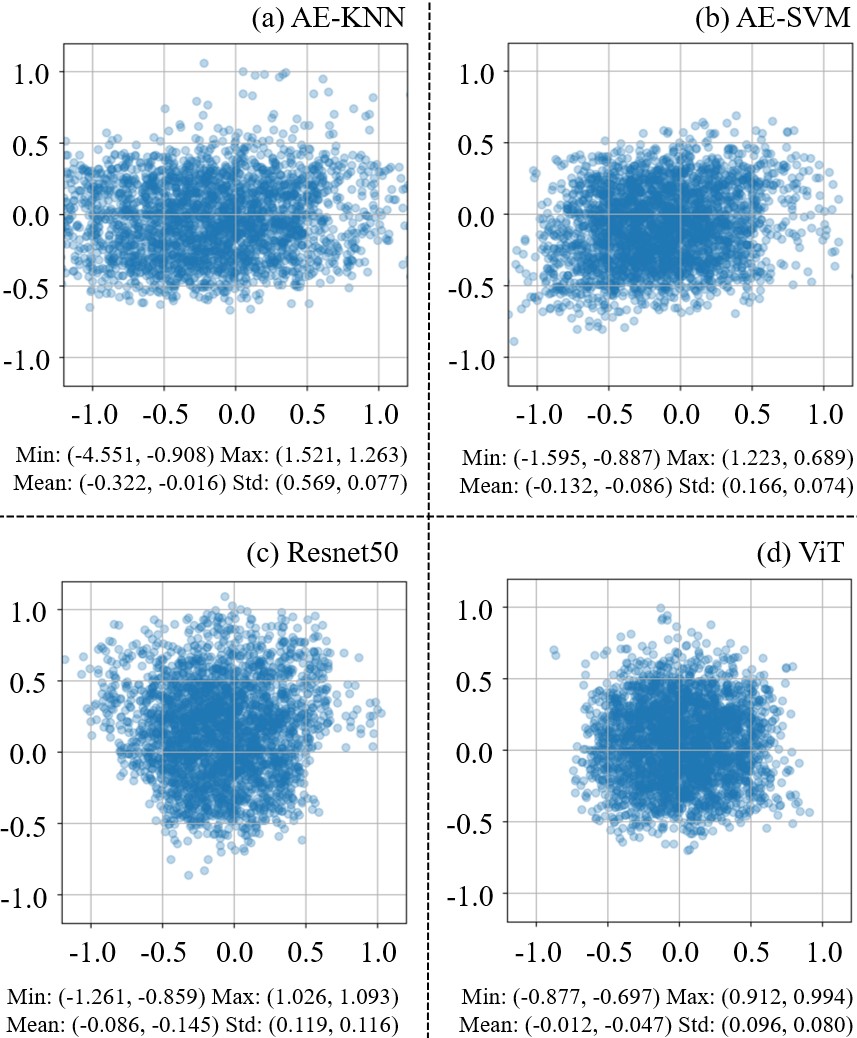}
    \caption{Results of the four classifiers against 3000 random deviations. Each point indicates a prediction error (different between predicted values and ground truth). The ViT method has the best performance. Metrics of axes: mm.}
    \label{fig:performance_main}
\end{figure}

Fig.\ref{fig:performance_main} shows the experimental results of 3000 random deviations. Each diagram in the figure illustrates the prediction error in a 2D plane. A (0, 0) prediction error means predicted deviations are the same as their correspondent ground truth. The results show that the ViT method has the best performance. Its prediction error is less than 1.0 mm in the plane.

\begin{figure}[!htbp]
    \centering
    \includegraphics[width=.97\linewidth]{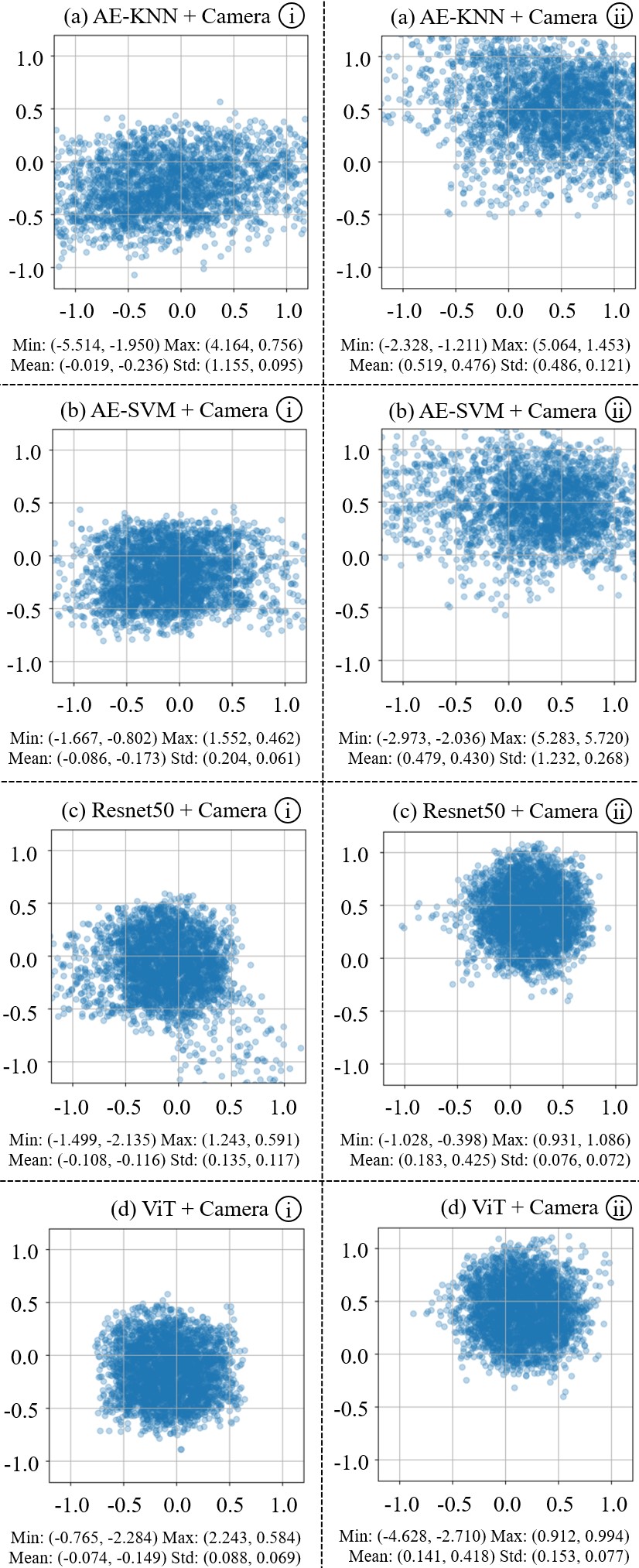}
    \caption{Results when a single camera is used. Left column: Camera \textcircled{\raisebox{-0.9pt}{ii}} is turned off; Right column: Camera \textcircled{\raisebox{-0.9pt}{i}} is turned off. Meanings of points and metrics are the same as Fig. \ref{fig:performance_main}.}
    \label{fig:performance_singlecam}
\end{figure}

We were particularly interested in the necessity of two cameras and thus carried out additional experiments by toggling on and off each camera. The results are shown in Fig. \ref{fig:performance_singlecam}. The left column of the figure is the results of the four methods only using the camera \textcircled{\raisebox{-0.9pt}{i}}. The right column are the results only using the camera \textcircled{\raisebox{-0.9pt}{ii}}. When using a single camera, the results exhibit a bias towards the camera's installation position. They are less precise than the results using both cameras (Fig. \ref{fig:performance_main}).

\subsubsection{Success rates of physical attachment} Physically attaching tips allow a certain deviation since the tips are unfixed and partially deformable. For this reason, we additionally conducted experiments to examine the success rates of physical attachment. We carried out experiments by placing the rack near the center of the aluminum base. The results for the four classifiers with different camera options are shown in Table \ref{tab:compare}. The success rates table section indicates that the physical attachment does not require a zero prediction error to ensure successful insertion. For example, the dual-camera Resnet50 and ViT classifiers could reach 100\% successful insertion rates, although their maximum prediction error reaches more than 1 mm. The table's average steps section shows the average number of correction steps before attaching tips. It corresponds to the second exception mentioned in Section V.E (more precisely, the average number of ``Counter'' shown in the purple dashed box of Fig. \ref{fig:exception}). The average steps indicate that the dual-camera Resnet50 and ViT classifiers do not lead to overshoot. However, using a single camera leads to overshoot and lower performance.

\begin{table}[htbp]
\centering
\caption{Results of physical attachment}
    \begin{tabular}{@{\extracolsep}clcccc}
    \toprule
    \multicolumn{1}{l}{} & & AE-KNN & AE-SVM& Resnet50 & ViT \\ 
    \midrule
    \multirow{3}{*}{\begin{tabular}[c]{@{}c@{}} Success rates\end{tabular}}
    & \begin{tabular}[c]{@{}l@{}} cam\textcircled{\raisebox{-0.9pt}{i}}\end{tabular} & 93.75\% & 87.5\% & 99.0\% & 92.7\%  \\ 
    & \begin{tabular}[c]{@{}l@{}} cam\textcircled{\raisebox{-0.9pt}{ii}}\end{tabular} & 92.7\% & 94.8\% & 99.0\% & 95.8\%  \\ 
    \cmidrule{2-6}
    \multirow{5}{*}{} & both & 96.9\% & 92.7\% & 100.0\% & 100.0\%  \\
    \midrule
    \multirow{3}{*}{\begin{tabular}[c]{@{}c@{}} Average steps\end{tabular}}
    & \begin{tabular}[c]{@{}l@{}} cam\textcircled{\raisebox{-0.9pt}{i}}\end{tabular} & 2.24 & 2.24& 1.31 & 2.02  \\ 
    & \begin{tabular}[c]{@{}l@{}} cam\textcircled{\raisebox{-0.9pt}{ii}}\end{tabular} & 1.89 & 1.65& 1.16 & 1.90  \\ 
    \cmidrule{2-6}
    \multirow{5}{*}{} & both & 1.52 & 1.19& 0.88 & 0.87 \\
    \bottomrule
    \end{tabular}
\label{tab:compare}
\end{table} 

Besides the above physical experiments, we also examined the influence of training data's rotation intervals. When acquiring the training data set, we set the rotation variation to be in the ranges [-30$^\circ$, 30$^\circ$] and [150$^\circ$, 210$^\circ$] with 5$^\circ$ interval (Section V.C). We chose 5$^\circ$ as it was dense enough considering our spiral triangulation method. However, we were interested in if a larger rotation interval still provides satisfying performance and how the rotation intervals would influence the classification and insertion results. Thus, we obtained new training data sets with rotation intervals of 10$^\circ$, 20$^\circ$, 30$^\circ$, and 45$^\circ$ by sampling the original data set. We used the new training data sets to train Resnet50 and ViT classifiers, respectively, and examined the performance of the newly trained classifiers to understand the influence of different rotational intervals. The results are shown in Table \ref{tab:intervals}. We did not conduct experiments with single cameras and the AE-KNN and AE-SVM classifiers since they had a less satisfying performance even with a 5$^\circ$ rotational interval. From the results, we can see that 5$^\circ$ is a good choice for both classifiers. It leads to high model accuracy, high success rates, and small average steps. As the interval increases, the performance drops significantly. There is no significant difference between the success rates and model accuracy of the ViT classifiers trained with 5$^\circ$ and 10$^\circ$ intervals (the first two rows of the ViT section). However, we could observe a longer average number of correction steps in the classifier using data with 10$^\circ$ intervals. The longer average steps mean the prediction results were less stable. The classifier might misjudge corrected results as non-zero classes again with high probability. The last three Resnet50 classifiers did not have success rates and average steps data because their trained models had less than 50\% accuracy and were ignored.

\begin{table}[htbp]
\renewcommand\arraystretch{1.2}
\centering
\caption{Results with different rotation intervals}
\begin{threeparttable}
    \begin{tabular}{@{\extracolsep}ccccccc}
    \toprule
    \multicolumn{1}{l}{} & Intervals & Succ. rates & Avg. steps & Model acc. & Dataset*\\ 
    \midrule
    \multirow{5}{*}{\begin{tabular}[c]{@{}c@{}} ViT\end{tabular}}
    & \begin{tabular}[c]{@{}l@{}}  5$^\circ$\end{tabular} & 100.0\% & 0.87 & 0.99 & 416  \\ 
    & \begin{tabular}[c]{@{}l@{}} 10$^\circ$\end{tabular} & 100.0\% & 1.80 & 0.99 & 208   \\ 
    & \begin{tabular}[c]{@{}l@{}} 20$^\circ$\end{tabular} & 100.0\% & 1.48 & 0.97 & 104   \\ 
    & \begin{tabular}[c]{@{}l@{}} 30$^\circ$\end{tabular} & 91.7\% & 2.76 & 0.94 & 69   \\ 
    & \begin{tabular}[c]{@{}l@{}} 45$^\circ$\end{tabular} & 86.5\% & 2.94 & 0.92 & 46  \\ 
    \midrule
    \multirow{5}{*}{\begin{tabular}[c]{@{}c@{}} Resnet50\end{tabular}}
    & \begin{tabular}[c]{@{}l@{}}  5$^\circ$\end{tabular} & 100.0\% & 0.88 & 0.96 & 416  \\ 
    & \begin{tabular}[c]{@{}l@{}} 10$^\circ$\end{tabular} & 58.3\% & 4.23 & 0.67 & 208\\
    & \begin{tabular}[c]{@{}l@{}} 20$^\circ$\end{tabular} & - & - & 0.38 & 104 \\ 
    & \begin{tabular}[c]{@{}l@{}} 30$^\circ$\end{tabular} & - & - & 0.22 & 69 \\ 
    & \begin{tabular}[c]{@{}l@{}} 45$^\circ$\end{tabular} & - & - & 0.13 & 46 \\ 
    \bottomrule
    \end{tabular}
    \begin{tablenotes}[flushleft]
    \item * The dataset column shows the number of training images in each class. Larger intervals lead to smaller training image sets. The smaller sets may also be an essential factor that impairs classification performance.
    \end{tablenotes}
\end{threeparttable}
\label{tab:intervals}
\end{table}

\subsection{Performance of Liquid Dispensing Tasks}

\begin{figure*}[!htbp]
    \centering
    \includegraphics[width=\linewidth]{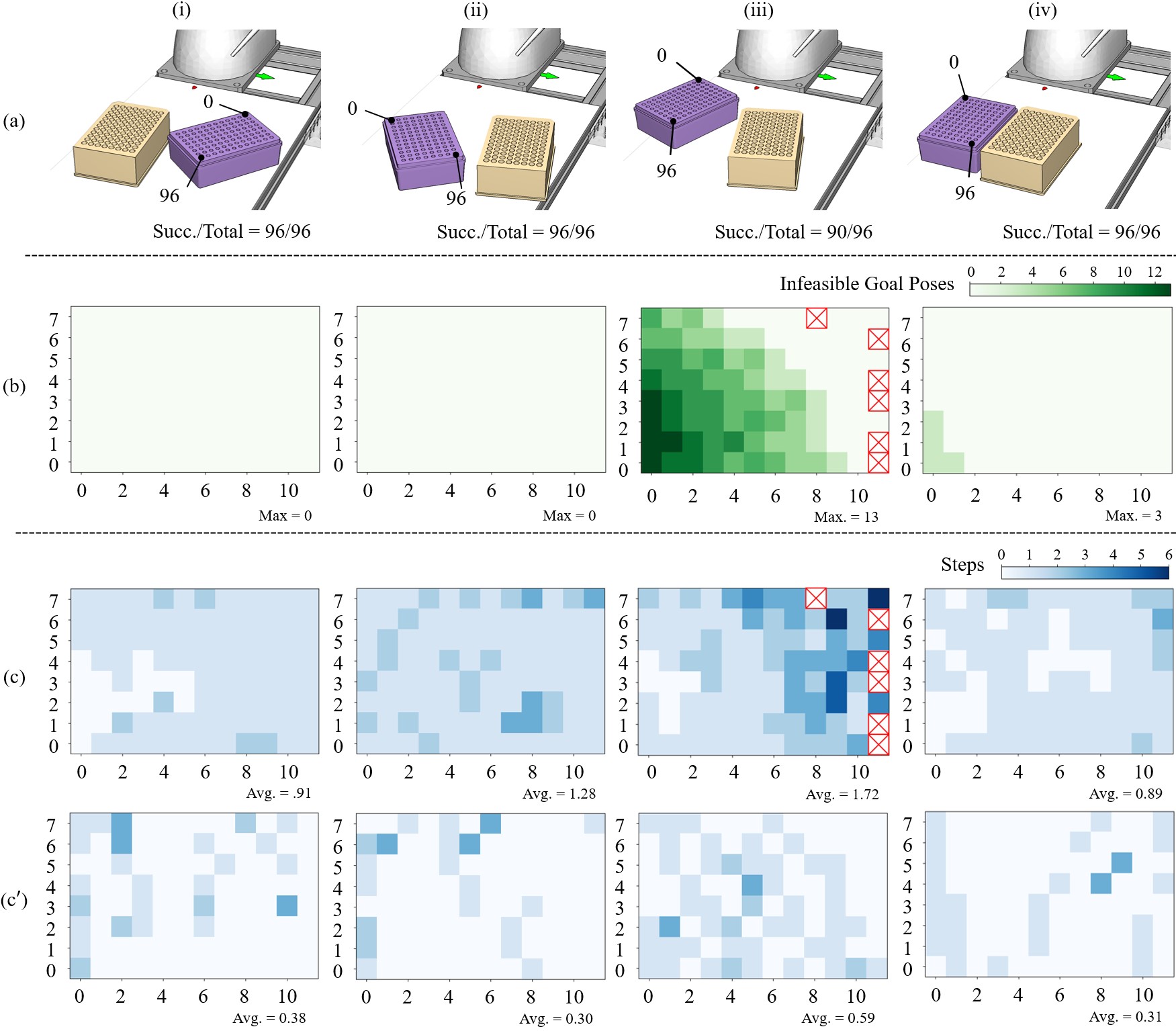}
    \caption{Results of real-world liquid dispensing tasks. Each column corresponds to a different placement. Row (a): Rack and plate poses obtained using direct teaching. Row (b): Number of infeasible goal poses encountered at each of the 96 tips during motion planning. Row (c): Number of correction steps at each tip before inserting the pipette shaft. Row (c$^\prime$): Results after including closed-loop pose update using previously predicted deviations.}
    \label{fig:randp}
\end{figure*}

We also examined the system's performance for completing a real-world liquid dispensing routine. We used the ViT classifier trained with 5$^\circ$ rotation intervals to conduct the experiments. Initially, we randomly placed the tip rack and microplate on the base of the robotic system and used the direct teaching method to obtain their poses. For each of the 96 tips in the hosting rack, we computed their coordinates based on the rack pose and got intermediate planning goals. Then, we planned the robot motion to move to the intermediate goals. The system took pictures, cropped and merged them into a single image, and classified the merged image to determine the correcting motion. After that, the robot performed the correcting motion, attached tips while considering the exceptions mentioned in Section V.E, and finished the aspiration, dispensing, and disposing motion. We observed the above process for every 96 tips and examined the success rate. We also simultaneously made a record of the two exceptions to compare different classification methods. The first exception was the number of rotation times before finding a reachable robot pose. The second one was the average number of re-classification / steps.

Fig. \ref{fig:randp} shows the results using four randomly placed rack and plate configurations. Row (a) of the figure illustrates the rack and plate poses obtained using direct teaching. The labels below each figure in the row show the number of successfully used tips. Row (b) shows the number of infeasible goal poses during motion planning. The row comprises four diagrams. Each shows the result of the correspondent rack poses in row (a). The four diagrams illustrate the number of infeasible goal poses for each of the tips in the rack as green gradient grids. The grids with darker green colors had a more significant number of infeasible poses. The grids with lighter green colors had a smaller number. In total, 96 tips are in the track; thus, each diagram comprises 96 gradient grids. The maximum number of infeasible goal poses for each rack pose are shown under respective diagrams. Row (c) shows each tip's correction steps (the ``Counter'' value in the purple dashed box of Fig. 14). Like row (b), we used gradient grids to represent the step number. Grids with darker blue colors imply that more correction steps were required. Grids with lighter blue colors imply fewer correction steps. The labels under the four diagrams in the row show the average number of correction steps for each pose.

The crossed boxes in the third diagrams of rows (b) and (c) denote the positions of failed tips\footnote{Only the third rack and plate configuration had failures. All tips in the other three configurations were successfully for fin liquid dispensing.}. We further studied the reason for these failures. We found that the significant absolute errors near the boundary of the robot's workspace caused the failures. The rack position in the third configuration was very near to the robot. When obtaining its position using the direct teaching method, the human teacher rotated the robot to reach the first tip, leading to a more significant absolute positioning error. On the other hand, the robot did not need to be rotated much for further tips, and the absolute positioning errors were minor. The different absolute positioning errors resulted in considerable uncertainty in the obtained rack pose. It led to more than 4.5$mm$ deviation and caused failures at the tips far away from ID 0 (tips with larger ID values). Besides the crossed boxes, we could also observe several very dark grids in the third diagram of the third row. The robot carried out more than three steps of correction at them. The reason was that the classifier misrecognized a deviation as a class of a nearby tip. The robot moved inversely and got more deviated.

We could solve the misrecognition problem by reducing the predicted deviation in a previous tip when determining the ensuing planning goals. The experimental results in Fig. \ref{fig:randp} were based on an open-loop implementation where the deviations at each tip were corrected independently without considering previously perceived deviations. Instead of this open-loop method, we considered closing the loop by updating the rack pose based on previous visual classification results. The closed-loop update could help avoid large deviations at the further tips and assure robust tip attachment and liquid dispensing. Row (c$^\prime$) of Fig.\ref{fig:randp} shows the results after including the closed-loop update. The performance (succ./total or avg. steps) improved significantly compared with the open-loop results shown in (c).

\subsection{Working with an Existing Automation System}

\begin{figure}[!htbp]
    \centering
    \includegraphics[width=.97\linewidth]{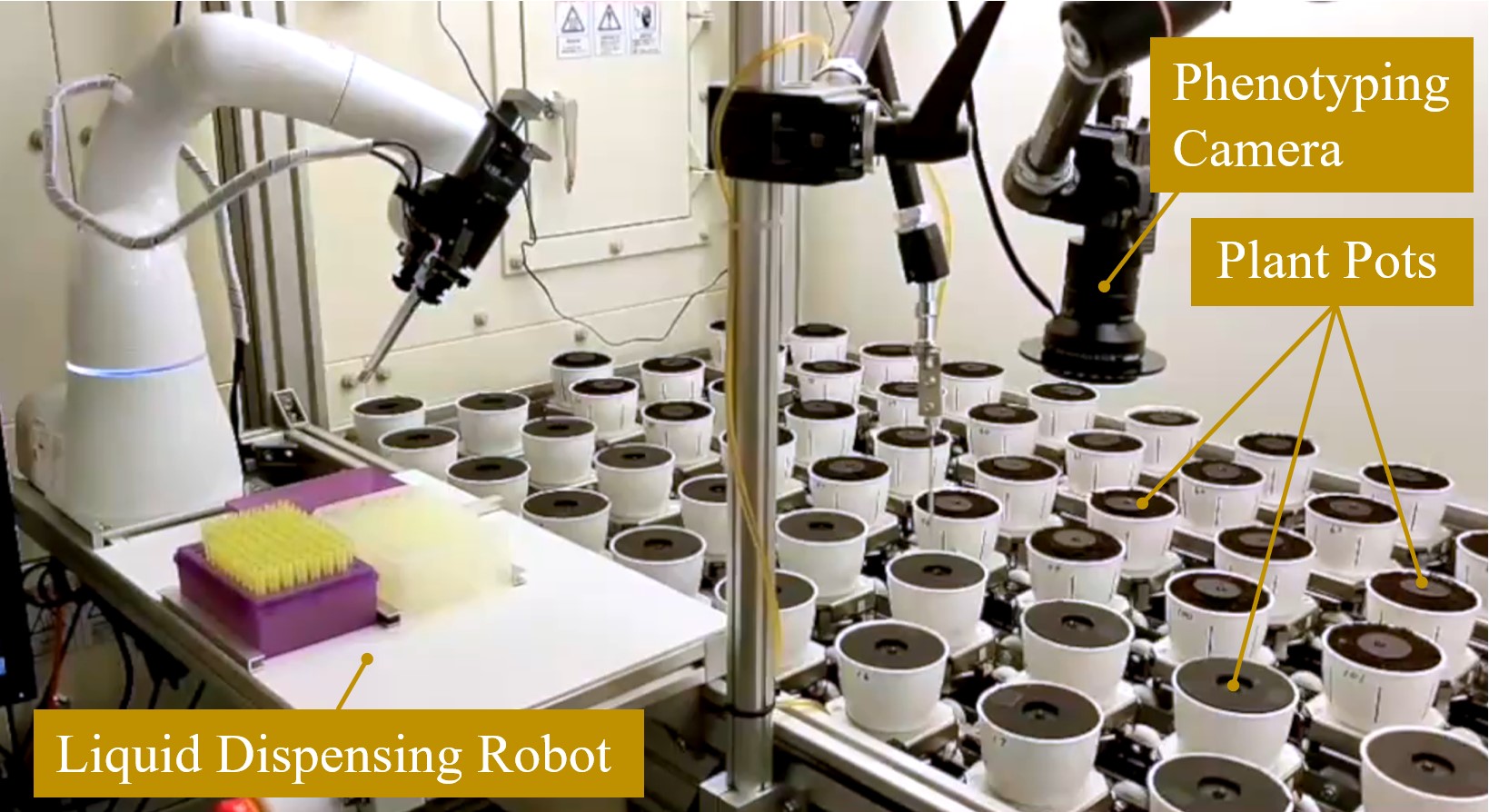}
    \caption{Deploying the developed robotic system for dispensing chemicals to plant pots transported by RIPPS.}
    \label{fig:ripps}
\end{figure}

The developed robotic system was deployed besides RIPPS \cite{fujita2018ripps} for dispensing chemicals. Fig. \ref{fig:ripps} shows a picture of the robot collaborating with the RIPPS' pot rotating mechanism. After cameras observed each pot in RIPPS, the system would dispatch a signal for the robot to begin dispensing liquid. The robot began to repeat the dispensing routine after receiving the signal. It used the classification and correction methods proposed above to attach tips sequentially, and followed a configurable protocol to determine the sequence for aspiring liquid from the micro-plate wells. The robot was able to handle the tips and liquid robustly with the proposed methods. Interested readers are suggested to see the supplementary video published with this manuscript for details.

\section{Conclusions and Future Work}

This paper integrated a manual pipette into a collaborative robot manipulator for liquid dispensing. We modified a parallel gripper to hold the manual pipette, took advantage of a collaborative robot manipulator's ``direct teaching'' mode to recognize randomly placed tip racks and plates, employed interleaved search and motion planning to generate robot motion, and applied vision-based classifiers to predict deviation and correct errors. Through various experiments, we confirmed that the developed system is robust for liquid handling tasks. It can flexibly deal with randomly placed labware and can be quickly deployed in a lab to work with existing automation devices for simultaneous liquid dispensing and analytical measurements.

In the future, we are interested in adapting the pipetting end-effector for other types of pipettes and generalizing the error correction methods to other tip types.

\bibliographystyle{IEEEtran}
\bibliography{citations.bib}

\end{document}